\documentclass{openmoss}

\graphicspath{{assets/}{./}}
\usepackage{helvet}

\usepackage{amsmath}
\usepackage{natbib}
\usepackage{graphicx}
\usepackage{subcaption}

\usepackage[utf8]{inputenc}
\usepackage[T1]{fontenc}
\usepackage{hyperref}
\usepackage{url}
\usepackage{booktabs}

\usepackage{amsfonts}
\usepackage{nicefrac}
\usepackage{microtype}
\usepackage{wrapfig}
\usepackage{amssymb}

\usepackage{titletoc}
\usepackage{minitoc}

\usepackage{array}
\usepackage{etoolbox}

\definecolor{lightblue}{RGB}{200, 230, 255}
\definecolor{headerblue}{RGB}{150, 200, 255}

\usepackage{pgfplots}
\pgfplotsset{compat=1.18}
\usepackage{xcolor}
\usepackage{float}
\usepackage{comment}
\usepackage{multirow}
\usepackage{makecell}
\usepackage{siunitx}
\usepackage{tikz}
\usepackage{pgf-pie}
\usepackage[export]{adjustbox}

\usepackage{ragged2e}
\usepackage{tabularx}
\usepackage{caption}
\usepackage{enumitem}
\usepackage{pifont}
\usepackage[hang,flushmargin]{footmisc}

\usepackage{tcolorbox}
\tcbuselibrary{breakable}
\tcbuselibrary{skins}

\usepackage{listings}

\usepackage{threeparttable}
\usepackage{longtable}
\usepackage{setspace}

\usepackage[scheme=plain,fontset=none]{ctex}
\usepackage{tabularx}

\newcolumntype{Y}{>{\raggedright\arraybackslash}X}
\newcolumntype{P}[1]{>{\raggedright\arraybackslash}p{#1}}

\newcolumntype{L}[1]{>{\RaggedRight\arraybackslash}p{#1}}
\newcolumntype{C}[1]{>{\centering\arraybackslash}p{#1}}
\DeclareUnicodeCharacter{00E9}{\'{e}}
\DeclareUnicodeCharacter{00F6}{\"{o}}
\DeclareUnicodeCharacter{0117}{\.{e}}
\DeclareUnicodeCharacter{017D}{\v{Z}}
\DeclareUnicodeCharacter{2013}{--}
\DeclareUnicodeCharacter{2014}{---}
\DeclareUnicodeCharacter{2018}{\textquoteleft}
\DeclareUnicodeCharacter{2019}{\textquoteright}

\usepackage[table]{xcolor}

\definecolor{oursgray}{gray}{0.95}

\definecolor{MossCyan}{HTML}{82D9FF}
\definecolor{MossBlue}{HTML}{82B1FF}

\definecolor{lightblue}{RGB}{200, 230, 255}
\definecolor{headerblue}{RGB}{150, 200, 255}
\definecolor{oursgray}{gray}{0.95}
\definecolor{MossCyan}{HTML}{82D9FF}
\definecolor{MossBlue}{HTML}{82B1FF}
\definecolor{tickG}{HTML}{00C853}
\definecolor{crossR}{HTML}{FF1744}
\definecolor{tickG}{HTML}{00C853}
\definecolor{crossR}{HTML}{FF1744}

\newtcolorbox{promptbox}[2][]{
    colback=white,
    coltext=black,
    arc=3mm,
    boxrule=0.5pt,
    colframe=black!60!white,
    title={#2},
    colbacktitle=black,
    coltitle=white,
    fonttitle=\bfseries,
    top=8pt,
    bottom=8pt,
    left=10pt,
    right=10pt,
    breakable,
    before upper={%
        \linespread{1}\selectfont
        \setlength{\parskip}{1ex plus 0.2ex minus 0.2ex}%
        \setlength{\parindent}{0pt}%
    },
    #1
}

\openmosslogo{teai}
\title{SWE-bench Science: Can Coding Agents Resolve Engineering Tasks in Science?}

\author{
Zhipeng Xu$^{1,2}$,
Jiahao Lu$^{1,2}$,
Yining Zheng$^{1,2}$,
Yuxin Wang$^{1,2, \dagger}$,
Xipeng Qiu$^{1,2,\dagger}$\\[2mm]
{\normalfont \normalsize $^{1}$Shanghai Innovation Institute},
{\normalfont \normalsize $^{2}$Fudan University}\\[1mm]
{\normalfont \normalsize $^{\dagger}$Corresponding author: wangyuxin@sii.edu.cn, 
xpqiu@sii.edu.cn}\\
{}
}

\abstract{
Software increasingly functions as part of the scientific instrument itself, making failures in scientific code capable of compromising not only program behavior but also the evidence underlying scientific conclusions. Yet existing evaluations of coding agents largely emphasize aggregate task success, providing limited insight into why agents fail when repairing scientific software. We introduce \textbf{SWE-bench Science}, a repository-level benchmark for scientific software engineering comprising 119 tasks from 98 GitHub repositories across 20 scientific domains. Each task is organized into one of three paradigms: Issue-driven, Expert-exploratory, and Engineering-integration. Even the best-performing agent, \textbf{Claude Code with Opus-5 (max), achieves a pass@1 below 50\%}, highlighting the substantial challenges posed by scientific software engineering. We identify four recurring failure mechanisms: deficits in scientific knowledge or abstraction, misguided exploration or surface-level repair, incomplete repair coverage or system integration, and failures to generalize scientific knowledge beyond observed cases in our analysis. We further conduct a paired ablation that removes explicit scientific guidance while preserving the repository and executable engineering context. The results show that scientific knowledge is not uniformly beneficial: well-grounded information can constrain repair and improve average performance and token efficiency, whereas poorly aligned guidance can induce anchoring and does not necessarily improve exact repair success. Together, SWE-bench Science provides a broad testbed for studying both the capabilities and failure mechanisms of coding agents in scientific software engineering.
}
\checkdata[Code]{\url{https://github.com/OpenMOSS/SWE-bench-Science}}
\checkdata[Data]{\url{https://huggingface.co/datasets/OpenMOSS-Team/SWE-bench-Science}}
\checkdata[Leaderboard]{\url{https://swescience.github.io}}
\begin{document}
\maketitle
\begingroup
\renewcommand{\thefootnote}{\fnsymbol{footnote}}
\setcounter{footnote}{0}
\footnotetext[2]{Corresponding author.}
\endgroup

\begin{figure}[h]
    \centering
    \includegraphics[width=0.96\linewidth]{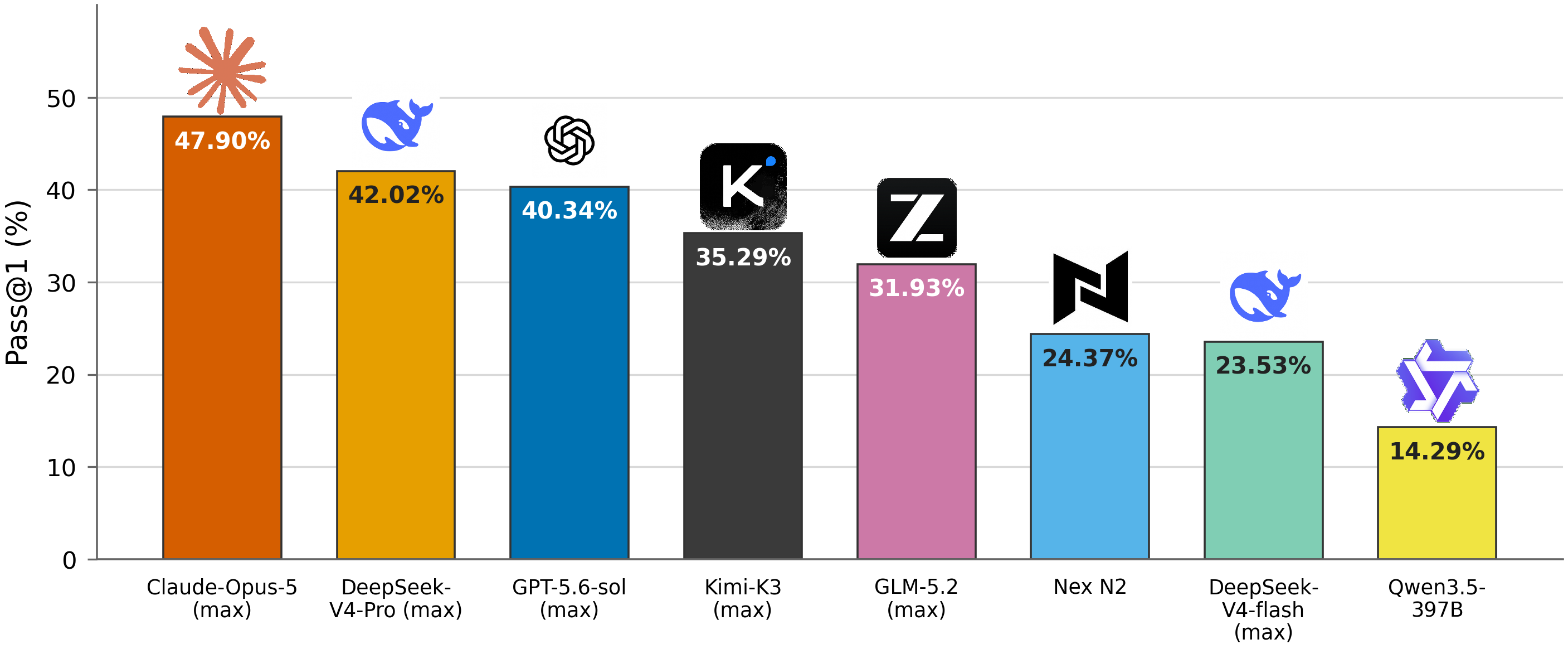}
    \caption{Pass@1 comparison of coding agents on SWE-bench Science.}
    \label{fig:model_pass_at_1_overview}
\end{figure}

\section{Introduction}

Software is no longer merely an aid to science; it is part of the instrument through which scientific claims are produced. Research groups rely on code to run simulations, process instrument outputs, train surrogate models, search chemical or biological candidates, manage high-throughput experiments, and reproduce published results. A defective patch can therefore corrupt not only a program output but also the evidence behind a scientific conclusion.

Understanding why coding agents fail on scientific software is as important as measuring whether their patches pass tests. A failed repair may reflect an incorrect scientific abstraction, superficial exploration of an observed symptom, incomplete integration across a software system, or an inability to generalize a scientific principle beyond visible cases. Aggregate test scores cannot distinguish these sources and therefore provide limited guidance for improving scientific coding agents. Existing benchmarks measure function synthesis and repository-level software repair~\cite{chen2021evaluating,austin2021program,jimenez2023swebench}, while scientific coding benchmarks cover curated research problems, paper-grounded implementations, scientific workflows, repository execution, and production scientific codebases~\cite{tian2024scicode,xia2026mmscicode,chen2025scienceagentbench,bogin2024super,bragg2026astabench,duston2025ainsteinbench}. However, repository-level coverage across scientific domains and the analysis of failure mechanisms remain limited, leaving broad cross-domain scientific software engineering underexplored.

We introduce \textbf{SWE-bench Science}, a repository-level benchmark containing 119 tasks from 98 unique GitHub repositories and spanning 20 scientific domains. Each task is manually inspected to verify its scientific contract, reproducibility, and evaluation validity. We organize the benchmark into three scientific task paradigms---\textbf{Issue-driven}, \textbf{Expert-exploratory}, and \textbf{Engineering-integration}---that capture complementary requirements in scientific software maintenance. Beyond measuring task success, we manually analyze unsuccessful attempts through four recurring scientific failure mechanisms: scientific-knowledge or abstraction deficits, misguided exploration or surface-level repair, incomplete repair coverage or system integration, and failures of scientific-knowledge generalization. Runtime or evaluation-path failures are recorded separately. We further construct a paired comparison that removes explicit scientific auxiliary information while preserving the repository and executable engineering context. The resulting analysis shows that scientific knowledge is not uniformly beneficial: its effect depends on how the knowledge is organized and grounded in executable evidence, because well-targeted information can constrain the repair while poorly aligned information can induce anchoring or substitute for independent validation.

Our contributions are:
\begin{enumerate}[leftmargin=*]
    \item We introduce a broad-coverage scientific software engineering benchmark with 119 tasks from 98 unique GitHub repositories, spanning 20 scientific domains, and evaluate representative frontier coding agents on these tasks.
    \item We analyze unsuccessful scientific software repairs and identify four recurring failure mechanisms, revealing concrete directions for improving scientific abstraction, disciplined exploration, system-wide repair, and scientific-knowledge generalization.
    \item We provide a paired scientific-knowledge ablation that removes explicit scientific guidance while preserving the executable engineering context. The results show that the organization and grounding of scientific knowledge matter in scientific software engineering: additional information can improve average scores and token efficiency, but does not automatically improve exact repair success.
\end{enumerate}

\section{Related Work}

\subsection{Code and Software Engineering Benchmarks}

Early code-generation benchmarks evaluate whether models can synthesize short functions from natural-language prompts. HumanEval~\cite{chen2021evaluating} and MBPP~\cite{austin2021program} are representative examples, focusing on unit-test correctness for self-contained programming problems. Later benchmarks increase realism by requiring repository-level modification and issue resolution. SWE-bench~\cite{jimenez2023swebench} evaluates whether language models can resolve real GitHub issues by generating patches against project test suites. More recent benchmarks extend this setting toward longer-horizon agentic engineering: SWE-Bench Pro targets complex enterprise-level repository tasks, DeepSWE uses original tasks and hand-written verifiers to reduce contamination and oracle ambiguity, and Terminal-Bench evaluates agents on hard, realistic tasks in command-line environments~\cite{deng2025swebenchpro,huang2026deepswe,merrill2026terminalbench}. Together, these benchmarks measure increasingly realistic software engineering capabilities, including repository inspection, fault localization, multi-file modification, tool use, and test-validated repair.

\subsection{LLMs for Scientific Workflows}

Scientific-code benchmarks mainly study two settings. SciCode and MMSciCode evaluate self-contained scientific programming and paper-grounded function implementation, respectively, making them closer to non-agentic programming evaluation than to repository-level software engineering~\cite{tian2024scicode,xia2026mmscicode}. A second line evaluates coding within broader research workflows: ResearchCodeBench and LMR-BENCH study research-paper code implementation and research-code reproduction~\cite{hua2025researchcodebench,yan2025lmrbench}; ScienceAgentBench, SUPER, and CSR-Bench emphasize data-driven scientific discovery, research-repository execution, and research-system deployment~\cite{chen2025scienceagentbench,bogin2024super,xiao2025csrbench}; and RExBench, AutoMat, and AstaBench evaluate research-code extension, scientific workflow recovery, and multi-stage scientific tasks~\cite{edwards2026rexbench,huang2026automat,bragg2026astabench}. AInsteinBench is closest to our setting because it targets scientific computing through issue-derived and synthesized feature tasks in production repositories~\cite{duston2025ainsteinbench}. SWE-Bench 5G studies specification-dependent telecom repair~\cite{chen2026swebench5g}. SWE-bench Science instead centers scientific software engineering as an applied engineering problem: coding agents must inspect production repositories, modify interacting components, and preserve scientific validity across \textbf{Issue-driven}, \textbf{Expert-exploratory}, and \textbf{Engineering-integration} tasks. This focus differs from solving standalone scientific programming problems or completing broader research workflows by evaluating the maintenance and extension of scientifically valid software systems.

\begin{table*}[htbp]
    \centering
    \footnotesize
    \begin{tabularx}{\textwidth}{l r Y r Y Y}
        \toprule
        \textbf{Benchmark} & \textbf{Task count} & \textbf{\# Scientific domains} & \textbf{GitHub repos} & \textbf{Scientific focus} & \textbf{Task source} \\
        \midrule
        SciCode~\cite{tian2024scicode} & 80 & 16 & -- & Function programming & expert \\
        MMSciCode~\cite{xia2026mmscicode} & 624 & 6 & -- & Function programming & expert \\
        ResearchCodeBench~\cite{hua2025researchcodebench} & 212 & 1 & 20 & Research code & expert \\
        LMR-BENCH~\cite{yan2025lmrbench} & 28 & 1 & 23 & Research code & expert \\
        ScienceAgentBench~\cite{chen2025scienceagentbench} & 102 & 4 & 30 & Research code & expert \\
        SUPER~\cite{bogin2024super} & 801 & -- & 694 & Research code & expert \\
        CSR-Bench~\cite{xiao2025csrbench} & 100 & 5 & 100 & Research code & -- \\
        RExBench~\cite{edwards2026rexbench} & 12 & -- & -- & Research code & expert \\
        AutoMat~\cite{huang2026automat} & 85 & 1 & -- & Research code & expert \\
        AstaBench~\cite{bragg2026astabench} & $>$2{,}400 & -- & -- & Research code & expert \\
        AInsteinBench~\cite{duston2025ainsteinbench} & 244 & 6 & 6 & Scientific computing & issue/expert \\
        SWE-Bench 5G~\cite{chen2026swebench5g} & 210 & 1 & 3 & Scientific software engineering & issue \\
        \textbf{SWE-bench Science (ours)} & 119  & 20  & 98 & Scientific software engineering & issue/expert/engineer \\
        \bottomrule
    \end{tabularx}
    \caption{Comparison of coding-agent benchmarks for science. SWE-bench Science covers 20 scientific domains.}
    \label{tab:related_comparison}
\end{table*}

\section{SWE-bench Science}

\begin{figure*}[t]
    \centering
    \includegraphics[width=\textwidth]{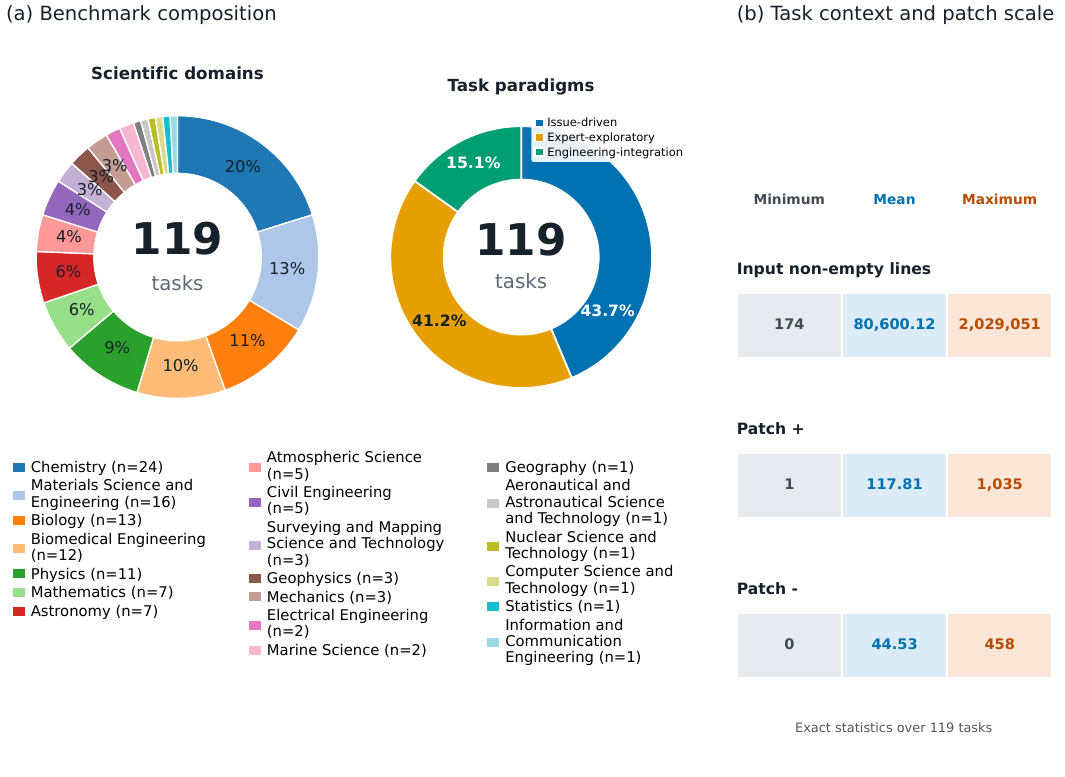}
    \caption{Benchmark coverage and scale over 119 tasks. (a) Scientific-domain distribution and task-paradigm composition. The domain legend reports the task count for each of the 20 domains; the task-paradigm chart contains 52 Issue-driven tasks (43.7\%), 49 Expert-exploratory tasks (41.2\%), and 18 Engineering-integration tasks (15.1\%). Chemistry is the largest domain with 24 tasks, followed by Materials Science and Engineering with 16, Biology with 13, Biomedical Engineering with 12, and Physics with 11. (b) Minimum, mean, and maximum input and reference-patch sizes, shown directly in horizontal summary bands. Input size is measured as non-empty physical lines in the task code, while Patch $+$ and Patch $-$ count additions and deletions in the standard reference diff.}
    \label{fig:discipline_distribution}
\end{figure*}

We cover 98 unique GitHub repositories and construct 119 tasks across 20 scientific domains, with the domain and task-paradigm composition shown in Figure~\ref{fig:discipline_distribution}(a). The five largest domains contain 76 tasks (63.9\%), while six domains contribute one task each. The task set contains 52 Issue-driven, 49 Expert-exploratory, and 18 Engineering-integration tasks. The complete scientific-domain counts are reported in Appendix Table~\ref{tab:dataset_inventory}. The benchmark also spans substantial variation in repository context and repair size. Figure~\ref{fig:discipline_distribution}(b) shows that non-empty input code averages 80,600.12 lines and ranges from 174 to 2,029,051 lines. Reference patches add 117.81 lines on average, with a range of 1--1,035, and delete 44.53 lines on average, with a range of 0--458. Direct labels preserve the exact values despite the large scale differences among these quantities.

\subsection{Task Schema}

The following fields are visible to the agent in the standard condition:
\begin{itemize}[leftmargin=*]
    \item \textbf{Repository snapshot}: the exact pre-task state with locked dependencies and runnable entry points, stripped of Git history, remotes, future changelogs, build caches, and solution-linked artifacts.
    \item \textbf{Problem statement}: a specification frozen before test and reference-patch authors inspect one another's artifacts.
    \item \textbf{Required scientific context \(c_i^{\mathrm{req}}\)}: local, versioned definitions and constraints required for a well-posed task; it is held fixed in every condition.
    \item \textbf{Public tests}: visible software and scientific assertions that support interactive debugging.
\end{itemize}

Evaluator-only fields include private tests, contract labels, reference and alternative-valid patches, scientific-rationale and localization support blocks, expected files, source and difficulty metadata, contamination tier, and post-hoc annotations. Private tests are mounted only after patch submission in a separate evaluator container. None of fields is available through the agent workspace, environment variables, logs, or task metadata.

\subsection{Example.}

\begin{figure*}[htbp]
    \centering
    \includegraphics[width=0.98\textwidth]{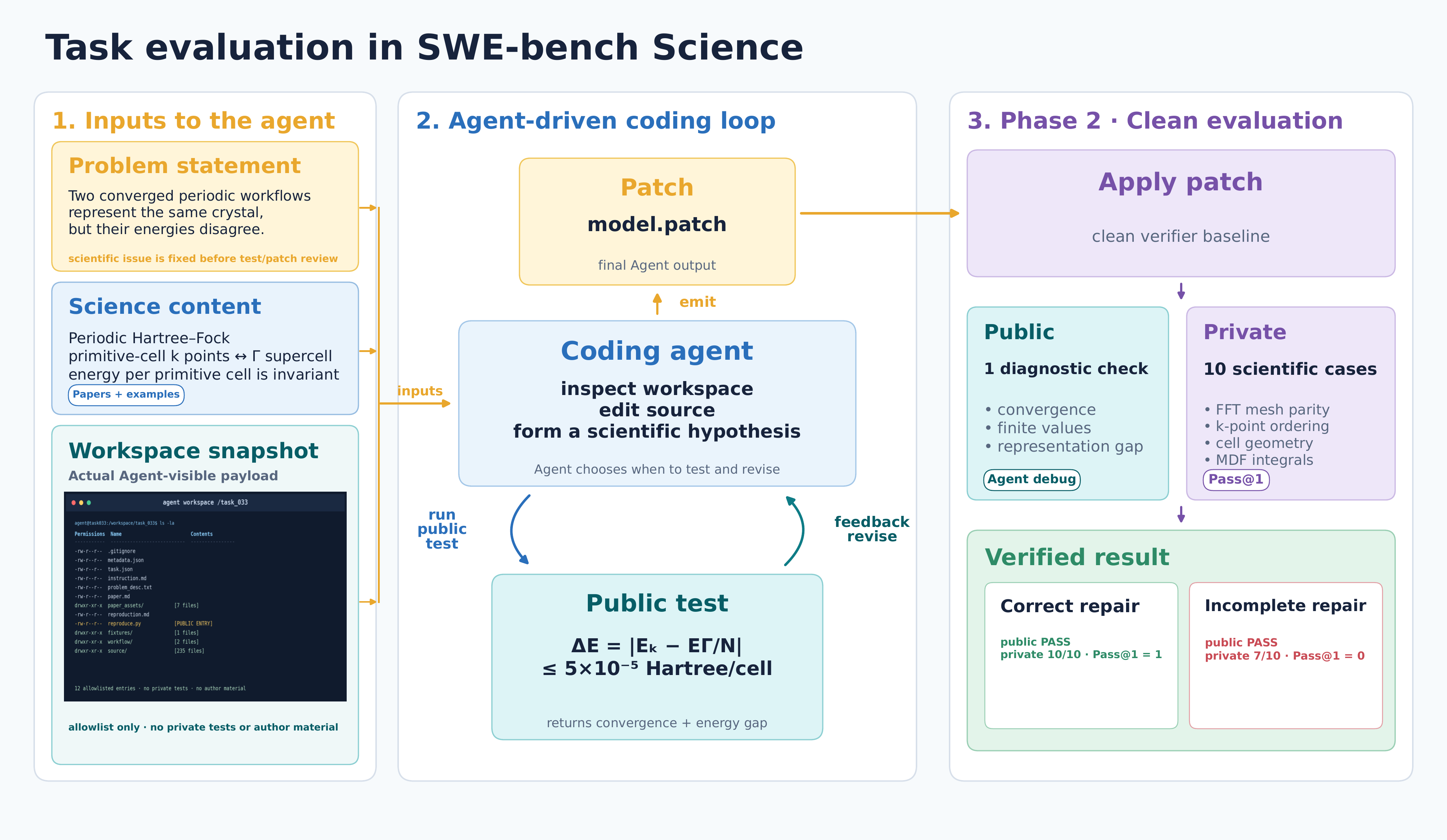}
    \caption{Evaluation pipeline in SWE-bench Science for an example task. The diagram shows the frozen agent-visible inputs, the agent-driven coding loop, and the clean evaluation phase with separate public diagnostics and private scientific cases.}
    \label{fig:task_example}
\end{figure*}

Figure~\ref{fig:task_example} illustrates an evaluation pipeline in the benchmark for an example task. The task asks the agent to reconcile two periodic representations of the same crystal while preserving their physical energy. The example shows the agent-visible problem statement, scientific context, and workspace snapshot as the input to the agent-driven coding loop and the clean evaluation phase. Public checks support debugging, whereas private scientific cases determine whether the submitted patch completes the intended repair.

\section{Benchmark Construction}

To evaluate agent performance across different
capability dimensions in real scientific software
development and maintenance, we establish a unified Chain-of-Evidence Protocol and
derive three task paradigms with clearly differentiated evaluation goals:
\textbf{Issue-driven}, \textbf{Expert-exploratory}, and
\textbf{Engineering-integration} tasks. Issue-driven tasks emphasize localized
repair and regression avoidance, Expert-exploratory tasks require autonomous
investigation of scientific discrepancies, and Engineering-integration tasks
evaluate cross-module completion of end-to-end scientific workflows.

\begin{figure}[htbp]
    \centering
    \includegraphics[width=0.8\textwidth]{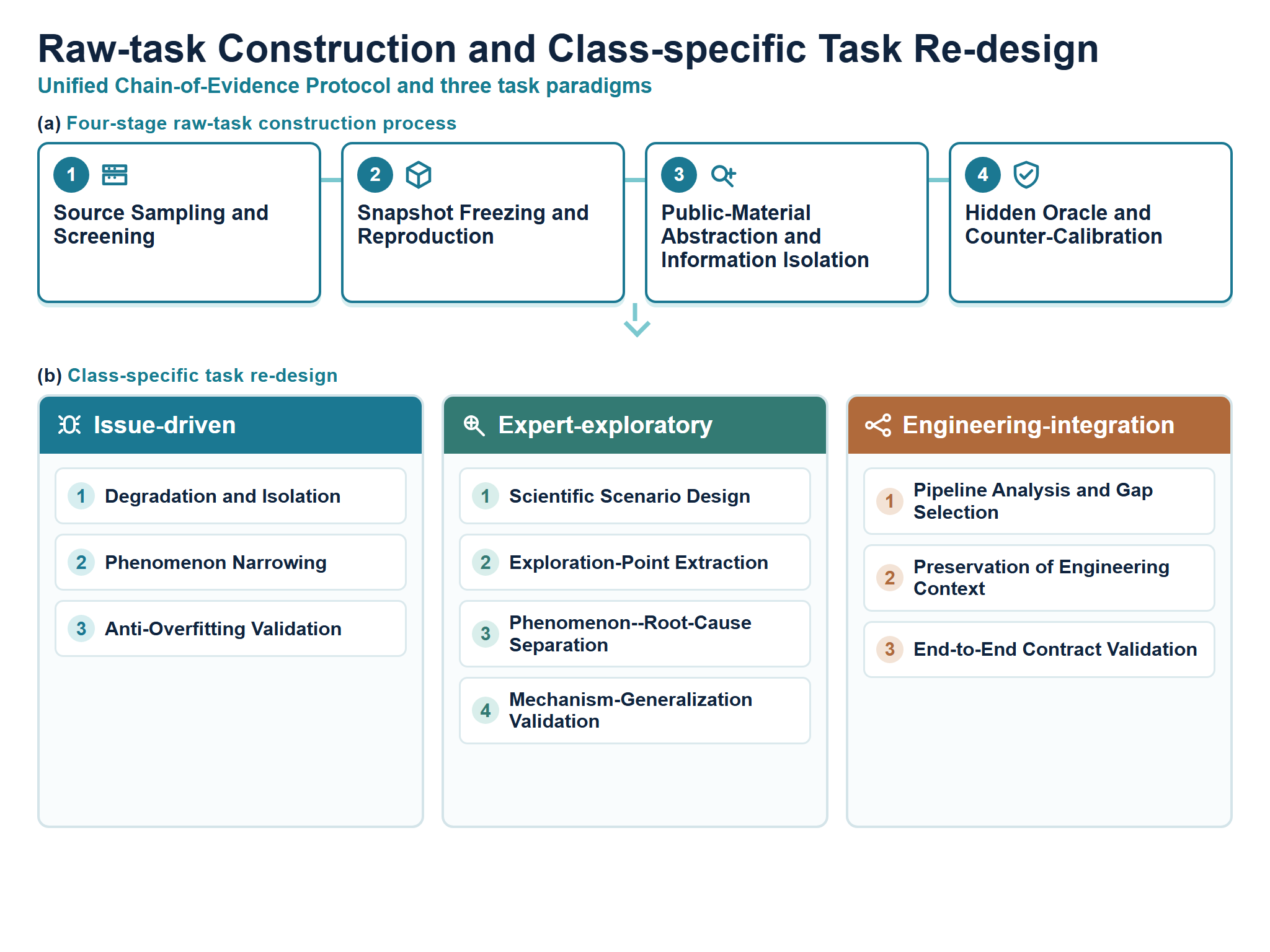}
    \caption{Combined overview of raw-task construction and class-specific task re-design. }
    \label{fig:benchmark_construction}
\end{figure}

All raw tasks are first generated according to a unified and auditable four-stage
construction process. The process is designed to control environmental
interference and prevent information leakage.

\begin{enumerate}[leftmargin=*]
    \item \textbf{Source Sampling and Screening.}

    We collect real issues, pull requests (PRs), commit records, and related
    literature from candidate open-source scientific-computing repositories.
    We then apply predefined screening criteria to remove overly simple fixes,
    tasks with unstable environment dependencies, tasks with severe solution
    leakage, and projects that substantially overlap with existing samples.

    \item \textbf{Snapshot Freezing and Reproduction.}

    We freeze the source-code snapshot immediately before the target defect or
    missing capability occurred, denoted by
    $\mathcal{S}_{\mathrm{bug}}$. The snapshot is executed in an isolated
    container to verify that the target anomaly or capability gap can be
    reproduced reliably. This step ensures that the observed failure is
    caused by the core algorithm or logical semantics rather than by
    environmental noise.

    \item \textbf{Public-Material Abstraction and Information Isolation.}

    Based on the intended behavioral contract, we extract the scientific
    invariants and data-flow constraints evaluated by each task. We construct
    a public evaluation package containing the public repository, a
    coarse-grained description of the observed phenomenon, a reproduction
    script, and the necessary domain-background documentation. The public
    materials do not reveal the patch location or hidden assertions.

    \item \textbf{Hidden Oracle and Counter-Calibration.}

    We construct hidden validation suites based on semantic equivalence and
    boundary conditions. The validators examine not only the execution
    results under normal inputs, but also include reverse checks for
    hard-coded solutions, heuristic pseudo-fixes, and incomplete repairs.
\end{enumerate}

Raw tasks are further re-designed according to their fit with the three task paradigms. Figure~\ref{fig:benchmark_construction} summarizes the four-stage raw-task construction process and the class-specific task re-design for the three paradigms.

\subsection{Class-Specific Task Re-design}
Through the differentiated construction strategies described below, we re-design the raw tasks to enable the benchmarking of three capabilities:

\begin{itemize}[leftmargin=*]
    \item \textbf{Issue-driven tasks} focus on repairing known defects;
    \item \textbf{Expert-exploratory tasks} focus on autonomously reasoning
    about unknown root causes in scientific scenarios;
    \item \textbf{Engineering-integration tasks} focus on understanding
    architecture across files and connecting a complete multi-module
    capability chain.
\end{itemize}

This task taxonomy avoids conflating different agent capabilities and
provides a structured experimental basis for evaluating large language
models in realistic scientific software engineering settings.

\subsubsection{Issue-Driven Tasks}

\paragraph{Definition and paradigm.}

Issue-driven tasks evaluate an agent's ability to repair real bugs under
low-distortion conditions. These tasks take a confirmed historical defect as
their direct definition starting point. Their central design principle is to
narrow the observable phenomenon while isolating the patch information.

\paragraph{Construction process.}

\begin{enumerate}[leftmargin=*]
    \item \textbf{Degradation and Isolation.}

    We analyze the historical issue and PR discussion, roll the source code
    back to the state before the problematic commit, and construct a minimal
    reproducible example (MRE).

    \item \textbf{Phenomenon Narrowing.}

    We reduce the upstream error to a coarse-grained and auditable phenomenon,
    such as inconsistent results across different computational paths or data
    loss under a particular format. The public materials remove any specific
    hints about the solution patch.

    \item \textbf{Anti-Overfitting Validation.}

    Hidden validators cover different data scales, boundary parameters, and
    input-order transformations. These validators test whether the repaired
    code genuinely restores the scientific semantics rather than merely
    fitting the public script.
\end{enumerate}

\subsubsection{Expert-Exploratory Tasks}

\paragraph{Definition and paradigm.}

Unlike Issue-driven tasks, which directly reproduce a known defect,
Expert-exploratory tasks simulate a real scientific exploration process.
Their definition starts from an exploration space containing a scientific
scenario. Some tasks may be inspired by historical issues or PRs, but those
issues or PRs do not serve as the definition starting point. These tasks
evaluate how an agent uses domain knowledge to perform autonomous black-box
or gray-box reasoning when the root cause of a complex phenomenon is unknown.

\paragraph{Construction process.}

\begin{enumerate}[leftmargin=*]
    \item \textbf{Scientific Scenario Design.}

    We first define a high-level scientific use case, such as molecular
    representation consistency calibration, measurement-chain bias analysis,
    medical-image coordinate alignment, or experimental-workflow
    verification.

    \item \textbf{Exploration-Point Extraction.}

    From theoretical materials, methodological literature, experimental
    phenomena, or discrepancies between results, we identify a core
    difference that is worth actively exploring. Upstream issues or PRs may
    provide optional background evidence, but they do not define the task.

    \item \textbf{Phenomenon--Root-Cause Separation.}

    We construct a runnable public workflow that preserves the relevant domain
    background and observable anomaly, such as abnormal convergence or biased
    results, while hiding the precise location of the defective source code.
    The agent must identify the root cause through autonomous observation,
    controlled comparisons, and scientific reasoning.

    \item \textbf{Mechanism-Generalization Validation.}

    Hidden validators change parameter scales, physical topologies, coordinate
    orderings, or boundary conditions. These tests determine whether the
    agent has genuinely inferred the scientific mechanism behind the
    observed phenomenon.
\end{enumerate}

\subsubsection{Engineering-Integration Tasks}

\paragraph{Definition and paradigm.}

Engineering-integration tasks go beyond single-point bug repair. They focus
on evaluating an agent's ability to understand system architecture in a
realistic and complex codebase, connect different modules, and complete a
full capability chain.

\paragraph{Construction process.}

\begin{enumerate}[leftmargin=*]
    \item \textbf{Pipeline Analysis and Gap Selection.}

    We analyze the repository's end-to-end call chain, including data loading,
    parameter interpretation, intermediate-representation (IR) construction,
    operator assembly, and numerical solving. We then select a functionality
    gap with an appropriate scope.

    \item \textbf{Preservation of Engineering Context.}

    The public source code preserves the complete real package structure and
    neighboring modules. Public reproduction scripts show state comparisons
    at multiple stages and identify where the functionality chain fails to
    connect correctly. This design requires the agent to navigate across files
    and develop a system-level understanding.

    \item \textbf{End-to-End Contract Validation.}

    Hidden validators include alternative execution paths, state-reset tests,
    and inter-module behavioral-contract tests. These validators ensure that
    the proposed repair achieves architecture-level integration across the
    entire module capability chain rather than merely making a single public
    test pass.
\end{enumerate}

\subsection{Scientific Auxiliary Information Separation}

The scientific-information factor measures the contribution of externally supplied domain knowledge, instantiated as scientific-rationale support. Of the 119 tasks, 91 allow this information to be separated from the supplied materials. For each eligible task, we compare paired conditions that differ only in auxiliary scientific information; the source-code snapshot, execution environment, public reproduction entry points, hidden validators, and required scientific context $c_i^{\mathrm{req}}$ remain fixed. Scientific auxiliary information denotes evidence relevant to scientific validity that is not directly available from source code or observable runtime behavior, including scientific rationales, upstream repairs, audit findings, paper excerpts, and expert guidance.

Withholding this information does not remove repository-intrinsic engineering cues, such as code structure, interfaces, traces, tests, and incomplete implementations, because doing so would change the software task itself. The ablation retains the executable context, task objective, observable symptoms, input data, and minimal interface documentation while removing scientific principles, equations or assumptions, expected properties, domain-specific diagnoses, and scientifically motivated repair strategies. It therefore estimates the marginal contribution of scientific auxiliary information given the same repository evidence, rather than performance in a clue-free setting. Results are analyzed in Section~\ref{analysis:science}.

\section{Experiments}

\paragraph{Coding Agents.} We evaluate eight agent configurations in our benchmark: GPT-5.6-sol with Codex at \textbf{max} reasoning depth, Claude-Opus-5 with Claude Code at \textbf{max}, Kimi-K3 with Kimi Code at \textbf{max}, GLM-5.2 with Codex at \textbf{max}, Nex N2 with Codex, DeepSeek-V4-flash-0731 with Claude Code at \textbf{max}, Qwen3.5-397B with Codex, and DeepSeek-V4-Pro-0813 with Claude Code at \textbf{max}. The comparison is shown in Table \ref{tab:mainresults}.

\paragraph{Evaluation.} We evaluate each submitted patch with public and private tests while keeping private tests and evaluator-only metadata outside the agent workspace. For each task--attempt, \textsc{PublicScore} is the mean score across all applicable public test cases, and \textsc{PrivateScore} is the corresponding mean across private test cases. \textsc{Fail2Pass} is the fraction of previously failing private tests that pass after the repair, while \textsc{Pass2Pass} is the fraction of previously passing private tests that remain passing; we set \textsc{Pass2Pass} to 1 when the previously passing set is empty. \textsc{Pass@1} is a binary exact-private-success measure that equals 1 only when every applicable private test passes. Table~\ref{tab:mainresults} reports task-level means for the public score, private score, Fail2Pass, Pass2Pass, and overall Pass@1, together with Pass@1 for Issue-driven, Expert-exploratory, and Engineering-integration tasks.

\subsection{Performance and Token Consumption}
\begin{table*}[htbp]
    \centering
    \footnotesize
    \setlength{\tabcolsep}{3pt}
    \begin{tabular}{llcccccccc}
        \toprule
        \multirow{2}{*}{\textbf{LLM}} &
        \multirow{2}{*}{\textbf{Harness}} &
        \multirow{2}{*}{\makecell{\textbf{Public}\\\textbf{score}}} &
        \multicolumn{3}{c}{\textbf{Private score}} &
        \multicolumn{4}{c}{\textbf{Pass@1}} \\
        \cmidrule(lr){4-6}\cmidrule(lr){7-10}
        & & & \textbf{Private score} & Fail2Pass & Pass2Pass & \textbf{Overall} & Issue & Expert & Engineering \\
        \midrule
        GPT-5.6-sol (max) & Codex & 98.32\% & \textbf{\underline{75.57\%}} & \textbf{\underline{69.98\%}} & 96.30\% & 40.34\% & 34.62\% & 46.94\% & 38.89\% \\
        Claude-Opus-5 (max) & Claude Code & 96.64\% & 75.11\% & 68.60\% & 97.37\% & \textbf{\underline{47.90\%}} & \textbf{\underline{38.46\%}} & \textbf{\underline{65.31\%}} & 27.78\% \\
        DeepSeek-V4-Pro (max) & Claude Code & \textbf{\underline{100.00\%}} & 73.16\% & 65.77\% & 96.58\% & 42.02\% & 26.92\% & 57.14\% & \textbf{\underline{44.44\%}} \\
        Kimi-K3 (max) & Kimi Code & 98.32\% & 66.34\% & 57.55\% & 94.94\% & 35.29\% & 25.00\% & 44.90\% & 38.89\% \\
        GLM-5.2 (max) & Codex & 94.12\% & 63.61\% & 53.81\% & \textbf{\underline{97.53\%}} & 31.93\% & 17.31\% & 46.94\% & 33.33\% \\
        Nex N2 & Codex & 93.28\% & 61.89\% & 51.09\% & 94.92\% & 24.37\% & 11.54\% & 36.73\% & 27.78\% \\
        DeepSeek-V4-flash (max) & Claude Code & 98.32\% & 61.41\% & 52.34\% & 95.74\% & 23.53\% & 19.23\% & 26.53\% & 27.78\% \\
        Qwen3.5-397B & Codex & 96.64\% & 51.79\% & 38.33\% & 95.16\% & 14.29\% & 5.77\% & 24.49\% & 11.11\% \\
        \bottomrule
    \end{tabular}
    \caption{Main experimental results. Scores are task-level means over the common 119 tasks. The best value in each column is bold and underlined; ties are retained.}
    \label{tab:mainresults}
\end{table*}
Table~\ref{tab:mainresults} shows that no single model leads every metric. DeepSeek-V4-Pro achieves the best public score and performs best on Engineering-integration tasks. GPT-5.6-sol leads the private score and \textsc{Fail2Pass}, indicating that it is the strongest at repairing failed private tests, while GLM-5.2 attains the best \textsc{Pass2Pass} at preserving tests that already pass. Claude-Opus-5 achieves the best overall \textsc{Pass@1} and leads on both Issue-driven and Expert-exploratory tasks. Even frontier models achieve Pass@1 scores below 50\%, underscoring the challenge of SWE-bench Science.

\begin{figure*}[htbp]
    \centering
    \includegraphics[width=0.98\linewidth]{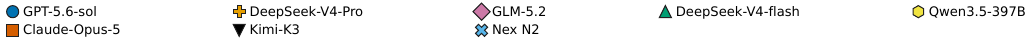}\\[-0.3em]
    \begin{subfigure}[t]{0.5\linewidth}
        \centering
        \includegraphics[width=\linewidth]{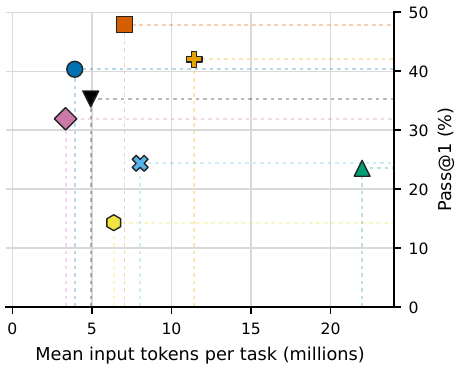}
        \caption{Mean input-token consumption.}
        \label{fig:pass_at_1_vs_input_tokens}
    \end{subfigure}%
    \begin{subfigure}[t]{0.5\linewidth}
        \centering
        \includegraphics[width=\linewidth]{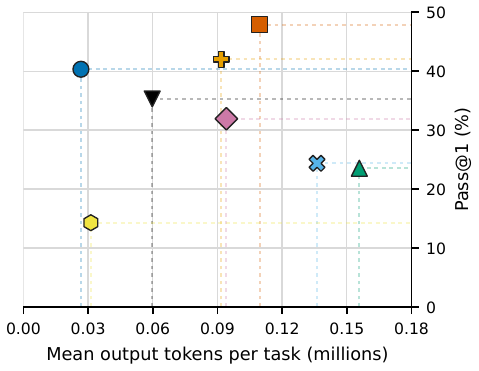}
        \caption{Mean output-token consumption.}
        \label{fig:pass_at_1_vs_output_tokens}
    \end{subfigure}
    \caption{Pass@1 versus mean token consumption per task over the common 119-task evaluation. Dashed guides connect each observation to the corresponding axes.}
    \label{fig:token_effect}
\end{figure*}

Figure~\ref{fig:token_effect} further shows the Pass@1 score versus token consumption for those agents. Claude-Opus-5 achieves the highest \textsc{Pass@1} with a moderate token budget, while GPT-5.6-sol reaches lower performance with much shorter outputs. DeepSeek-V4-Pro uses more input tokens and ranks between them. Kimi-K3 and GLM-5.2 achieves lower performances but consume fewer input tokens. For smaller models under 400B, Nex-N2 achieves best performance-token balance, better than DeepSeek-V4-flash and Qwen3.5-397B. The main difference therefore lies in how effectively each model--harness configuration uses its context and generation budget, rather than in token volume alone.

\section{Analysis}

\subsection{Observed Failure Mechanisms}

\begin{table*}[htbp]
    \centering
   
    \label{tab:error_attribution}
    \small
    \setlength{\tabcolsep}{4pt}
    \begin{tabular}{llccccc}
        \toprule
        \textbf{LLM} & \textbf{Harness} & \makecell{\textbf{Total}\\\textbf{errors}} & \makecell{\textbf{Knowledge/}\\\textbf{abstraction}} & \makecell{\textbf{Exploration/}\\\textbf{surface repair}} & \makecell{\textbf{Repair coverage/}\\\textbf{system integration}} & \makecell{\textbf{Scientific}\\\textbf{generalization}} \\
        \midrule
        GPT-5.6-sol (max) & Codex & 71 & 18 & 13 & 21 & 19 \\
        Claude-Opus-5 (max) & Claude Code & \textbf{\underline{58 (+4)}} & 24 & \textbf{\underline{2}} & 21 & 11 \\
        DeepSeek-V4-Pro (max) & Claude Code & 69 & \textbf{\underline{15}} & 12 & \textbf{\underline{19}} & 23 \\
        Kimi-K3 (max) & Kimi Code & 77 & 20 & 14 & 22 & 21 \\
        GLM-5.2 (max) & Codex & 81 & 20 & 14 & 24 & 23 \\
        Nex N2 & Codex & 90 & 31 & 14 & 23 & 22 \\
        DeepSeek-V4-flash (max) & Claude Code & 91 & 23 & 14 & 48 & \textbf{\underline{6}} \\
        Qwen3.5-397B & Codex & 102 & 26 & 15 & 33 & 28 \\
        
        \bottomrule
    \end{tabular}
     \caption{Error-count breakdown by coding agents over the common 119 tasks. The total column records attempts assigned to the four mutually exclusive scientific failure mechanisms defined in the text; for Claude-Opus-5, the parenthesized $+4$ denotes additional runtime or evaluation-path failures. The four category counts sum to the main total for each row. The lowest count in each column is bold and underlined.}
\end{table*}

We use four recurring scientific failure mechanisms in the audit reports. \textbf{Scientific-knowledge or abstraction deficit} denotes a repair based on an incorrect or incomplete scientific object, mathematical definition, or domain abstraction. \textbf{Misguided exploration or surface-level repair} denotes a patch that addresses the visible symptom or public metric without tracing the failure to an independent oracle or the underlying scientific contract. \textbf{Incomplete repair coverage or system integration} denotes a locally plausible repair that does not satisfy the requirements of the full software system: for example, one module is corrected while its interactions, data flow, shared invariants, or compatibility with other modules remain unpreserved. \textbf{Failure of scientific-knowledge generalization} denotes a repair that handles the observed scientific case but does not extend the same scientific principle to unseen conditions, equivalent representations, boundary regimes, or other variants that require scientific generalization. Claude-Opus-5 additionally has four \textbf{runtime or evaluation-path failures}, in which an attempt does not complete the intended execution or evaluation path and therefore cannot be attributed to one of the four scientific mechanisms.

Claude-Opus-5 produces the lowest categorized scientific-error count (58, plus 4 runtime or evaluation-path failures) and the fewest misguided-exploration or surface-level-repair errors (2). DeepSeek-V4-flash (max) records the fewest scientific-knowledge generalization errors (6), while DeepSeek-V4-Pro records both the fewest scientific-knowledge or abstraction errors (15) and the fewest incomplete-repair or system-integration errors (19).

\subsection{How Scientific Information Affects Performance}
\label{analysis:science}
\begin{figure}[htbp]
    \centering
    \includegraphics[width=0.7\columnwidth]{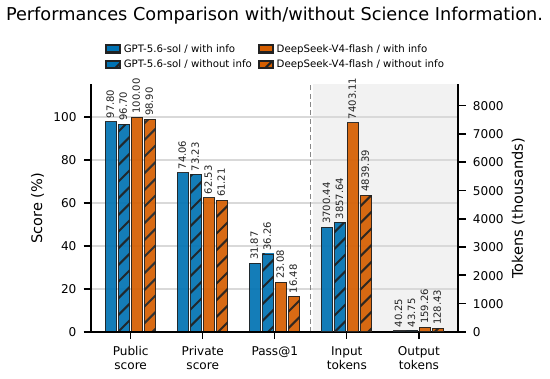}
    \caption{Scores and token consumption over the 91 tasks whose scientific-information content differs between conditions. GPT-5.6-sol and DeepSeek-V4-flash are each shown with and without scientific information on the same task subset. }
    \label{fig:scientific_information_ablation}
\end{figure}

On the 91-task separable subset, we compare GPT-5.6-sol (xhigh) with DeepSeek-V4-flash (high), each with and without scientific information. Figure~\ref{fig:scientific_information_ablation} summarizes their mean scores and token use.

For GPT-5.6-sol, scientific information slightly increases the mean public and private scores from 96.70\% and 73.23\% to 97.80\% and 74.06\%, but decreases Pass@1 from 36.26\% to 31.87\%. Mean input and output tokens also decrease from 3.86 million and 43.75 thousand to 3.70 million and 40.25 thousand. For DeepSeek-V4-flash, the same intervention raises the public score, private score, and Pass@1 from 98.90\%, 61.21\%, and 16.48\% to 100.00\%, 62.53\%, and 23.08\%, while increasing input and output tokens from 4.84 million and 128.43 thousand to 7.40 million and 159.26 thousand. These paired descriptive differences show model-specific score--cost responses but do not establish statistical significance or a causal effect.

\begin{figure*}[htbp]
    \centering
    \textbf{Pass@1 overlap on the 91-task separable subset}\\[-0.2em]
    \begin{subfigure}[t]{0.5\linewidth}
        \centering
        \includegraphics[width=\linewidth]{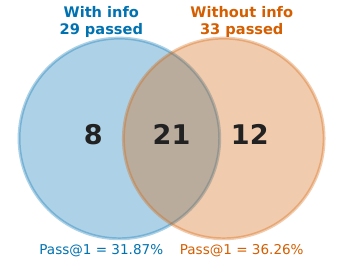}
        \caption{GPT-5.6-sol.}
        \label{fig:pass_at_1_overlap_gpt}
    \end{subfigure}%
    \begin{subfigure}[t]{0.5\linewidth}
        \centering
        \includegraphics[width=\linewidth]{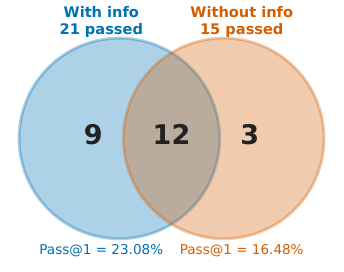}
        \caption{DeepSeek-V4-flash.}
        \label{fig:pass_at_1_overlap_deepseek}
    \end{subfigure}
    \caption{Task-level overlap of Pass@1 success on the 91-task separable subset. (a) GPT-5.6-sol passes 29 tasks with scientific information and 33 tasks without it; 21 pass in both conditions, 8 only with information, and 12 only without it. (b) DeepSeek-V4-flash passes 21 tasks with scientific information and 15 tasks without it; 12 pass in both conditions, 9 only with information, and 3 only without it. }
    \label{fig:pass_at_1_overlap}
\end{figure*}

Figure~\ref{fig:pass_at_1_overlap} shows how these aggregate changes arise from task-level transitions. GPT-5.6-sol solves eight tasks only with scientific information and twelve only without it, whereas DeepSeek-V4-flash solves nine only with information and three only without it. Inspection of the GPT-5.6-sol cases suggests that scientific information can supply semantic constraints absent from local code symptoms, such as limiting cases, coordinate consistency, independent observables, and authoritative interfaces. However, it can also induce anchoring, scope spillover, or premature reliance on a supplied explanation, underscoring the need for executable evidence and independent validation.

Within this two-model comparison, scientific information produces a larger Pass@1 gain for the lower-baseline DeepSeek-V4-flash configuration than for GPT-5.6-sol. This pattern suggests that weaker-performing models may benefit more from external scientific guidance, whereas stronger models may rely less on it.

\section{Limitations}

The number of tasks in each scientific domain is still relatively limited, which may reduce the reliability of comparisons across domains. In addition, the analysis of the role of scientific knowledge in SWE tasks remains relatively preliminary, with insufficient exploration of how domain-specific knowledge is actually used and how to make it contribute to successful task completion.

\section{Conclusion}

We introduced \textbf{SWE-bench Science}, a repository-level benchmark for evaluating whether coding agents can repair scientific software while preserving its scientific contracts. The benchmark contains 119 tasks from 98 GitHub repositories across 20 scientific domains and organizes them into \textbf{Issue-driven}, \textbf{Expert-exploratory}, and \textbf{Engineering-integration} paradigms. Its Chain-of-Evidence Protocol combines separated public and private tests with metrics for repair progress, exact success, and regression preservation, making it possible to distinguish visible-test performance from complete private-test correctness.

Across eight coding-agent configurations, the strongest result is 47.90\% \textsc{Pass@1}, while the same configuration attains a 96.64\% public score. This gap, together with the four recurring scientific failure mechanisms identified in unsuccessful repairs, shows that repository-level scientific software engineering requires scientific abstraction, disciplined exploration, system-wide integration, and generalization beyond visible cases. Paired comparisons on the 91-task scientific-information subset further reveal model-dependent effects: auxiliary scientific information lowers GPT-5.6-sol \textsc{Pass@1} from 36.26\% to 31.87\% but raises DeepSeek-V4-flash from 16.48\% to 23.08\%, with opposite changes in token use. These findings indicate that supplying scientific information alone does not guarantee a better repair; its value depends on how effectively an agent connects that information to repository evidence and validates it through execution. SWE-bench Science provides a foundation for measuring this capability, while broader model coverage is needed to understand how scientific information affects agent performance on scientific software engineering tasks.

\clearpage 
\bibliographystyle{unsrtnat}
\bibliography{main}
\appendix
\newpage
\section{Scientific Domain Coverage}
\label{appendix:datasetcoverage}

\begin{longtable}{p{0.78\textwidth}r}
\caption{Distribution of the 119 SWE-bench Science tasks across 20 scientific domains.}\label{tab:dataset_inventory}\\
\toprule
\textbf{Scientific domain} & \textbf{Tasks} \\
\midrule
\endfirsthead
\multicolumn{2}{c}{\tablename\ \thetable{} -- continued from previous page} \\
\toprule
\textbf{Scientific domain} & \textbf{Tasks} \\
\midrule
\endhead
\midrule
\multicolumn{2}{r}{Continued on next page} \\
\endfoot
\bottomrule
\endlastfoot
Chemistry & 24 \\
Materials Science and Engineering & 16 \\
Biology & 13 \\
Biomedical Engineering & 12 \\
Physics & 11 \\
Mathematics & 7 \\
Astronomy & 7 \\
Atmospheric Science & 5 \\
Civil Engineering & 5 \\
Surveying and Mapping Science and Technology & 3 \\
Geophysics & 3 \\
Mechanics & 3 \\
Electrical Engineering & 2 \\
Marine Science & 2 \\
Geography & 1 \\
Aeronautical and Astronautical Science and Technology & 1 \\
Nuclear Science and Technology & 1 \\
Computer Science and Technology & 1 \\
Statistics & 1 \\
Information and Communication Engineering & 1 \\
\midrule
\textbf{Total} & \textbf{119} \\
\end{longtable}

\section{Task-Level Repository and Knowledge Inventory}
\label{appendix:taskinventory}

This appendix provides the complete task-level annotation used to construct the benchmark. Each row links a task identifier to one of the 20 scientific domains used throughout the paper, together with its upstream repository, issue or pull request (when available), knowledge-domain scope, and scientific focus. The table contains all 119 tasks (IDs 001--119).

\begingroup
\scriptsize
\setlength{\tabcolsep}{2pt}
\renewcommand{\arraystretch}{1.12}
\begin{longtable}{L{0.045\textwidth}L{0.115\textwidth}L{0.145\textwidth}L{0.115\textwidth}L{0.255\textwidth}L{0.275\textwidth}}
\caption{Task-level scientific-domain, provenance, and scientific-knowledge inventory. A missing issue or pull-request reference indicates that no reference was provided in the source annotation.}\label{tab:task_knowledge_inventory}\\
\toprule
\textbf{Task} & \textbf{Scientific domain} & \textbf{Upstream repository} & \textbf{Issue/PR} & \textbf{Knowledge scope} & \textbf{Task focus} \\
\midrule
\endfirsthead
\multicolumn{6}{c}{\tablename\ \thetable{} -- continued from previous page} \\
\toprule
\textbf{Task} & \textbf{Scientific domain} & \textbf{Upstream repository} & \textbf{Issue/PR} & \textbf{Knowledge scope} & \textbf{Task focus} \\
\midrule
\endhead
\midrule
\multicolumn{6}{r}{Continued on next page} \\
\endfoot
\bottomrule
\endlastfoot
001 & Chemistry & \href{https://github.com/Auto-Mech/autochem}{\nolinkurl{Auto-Mech/autochem}} & PR \#637 & TS linear segment cap selection, rotational symmetry number, reaction edge relabeling invariance, rotor observation & The rotational symmetry of the TS graph must be determined by the real linear segment and reaction edge topology; changing the atom number or writing it in reverse cannot change the rotor/symmetry observation. \\
002 & Chemistry & \href{https://github.com/aditisingh4812/simplified_tuned_range_separated}{\nolinkurl{aditisingh4812/simplified_tuned_range_separated}} & Not specified & range-separated tuning, local density structure, bohr-based grid scale and size consistency & In range-separated tuning, the units of the density cutoff are not converted together with the grid representation, and the local thresholds are therefore misaligned. \\
003 & Materials Science and Engineering & \href{https://github.com/deepmodeling/dpdata}{\nolinkurl{deepmodeling/dpdata}} & Issue \#973 / PR \#983 & stress/virial tag preservation, ASE compatibility, read and write consistency in extxyz/QUIP-GAP parsing & extxyz parsing must convert stress into virial and be compatible with tensor components and field aliases, otherwise the stress semantics will be distorted. \\
004 & Biology & \href{https://github.com/arq5x/bedtools2}{\nolinkurl{arq5x/bedtools2}} & Issue \#750 & split-read exon overlap filtering, interval intersection semantics, hit determination on splicing boundaries & For split-read interval intersection, block overlap should be accumulated first and then fraction gate should be performed. Each block cannot be treated as an independent hit. \\
005 & Biomedical Engineering & \href{https://github.com/schorschinho/osprey}{\nolinkurl{schorschinho/osprey}} & Issue \#750 / PR \#751 & edited MRS subspectrum polarity correction, generalization from two subspectra to four subspectra, phase-cycle control propagation & Edited MRS requires polarity correction and phase-cycle control to be propagated through every subspectrum path; the fix cannot be confined to a single loader. \\
006 & Biology & \href{https://github.com/arq5x/bedtools2}{\nolinkurl{arq5x/bedtools2}} & Issue \#954 / PR \#959 & Query database tail count after EOF, same-chrom / later-chrom tails, Fisher right tail p-value consistency & Fisher statistics require full database counts, and query EOF cannot truncate chromosome sweeps early. \\
007 & Biomedical Engineering & \href{https://github.com/openmrslab/suspect}{\nolinkurl{openmrslab/suspect}} & Issue \#33 / PR \#34 & spectral registration initial value symbol, warm-start convention drift, frequency shift and phase shift consistency & The drift sign of spectral registration must be consistent with the optimization residual agreement, otherwise the compensation direction will be reversed. \\
008 & Physics & \href{https://github.com/PlasmaControl/DESC}{\nolinkurl{PlasmaControl/DESC}} & Not specified & first usable vacuum-plus-wall capability chain, outer-region operator payload, pseudo-vacuum metrics & In the vacuum-wall link, the outer-region geometry and coefficients must be transferred together, and different wall shapes cannot be pressed into the same diagnosis. \\
009 & Physics & \href{https://github.com/PlasmaControl/DESC}{\nolinkurl{PlasmaControl/DESC}} & Not specified & projection-wall geometry construction, external-region path, downstream reduced payload propagation & The projection wall cannot degenerate into a scaled wall, and the boundary derivatives and geometric parameters must be entered into the outer-region calculation together. \\
010 & Physics & \href{https://github.com/PlasmaControl/DESC}{\nolinkurl{PlasmaControl/DESC}} & Not specified & alternate perturbation-representation support, FOURIN, packed trig lookup, matrix assembly & The PARFAC-driven alternate perturbation representation must not fall back to the baseline; parity, phase, and coefficient assembly must all follow the alternate representation. \\
011 & Physics & \href{https://github.com/PlasmaControl/DESC}{\nolinkurl{PlasmaControl/DESC}} & Not specified & response-gated sideband capability, localized exchange propagation, energy diagnostics & Sideband response must propagate through the support domain, coefficient assembly, and energy diagnostics; changing only a local cutoff is insufficient. \\
012 & Physics & \href{https://github.com/PlasmaControl/DESC}{\nolinkurl{PlasmaControl/DESC}} & Not specified & 3-D cross-spectral coupling, coefficient assembly, energy channel, mode-family generalization & 3-D mode coupling must retain the effective support of the difference/sum two branches, and cross-modal coupling cannot be collapsed into a unified weight. \\
013 & Physics & \href{https://github.com/PlasmaControl/DESC}{\nolinkurl{PlasmaControl/DESC}} & Not specified & free / soft-fixed / hard-fixed boundary regimes, operators and energy closure ladder & The boundary response should be closed to coefficients and energies respectively by interval and cannot be smoothed out by global averaging. \\
014 & Physics & \href{https://github.com/PlasmaControl/DESC}{\nolinkurl{PlasmaControl/DESC}} & Not specified & prescribed wall geometry, vacuum-plus-wall chain, outer-region operator payload, reduced response blocks & The TERPSICHORE link must be closed end-to-end and cannot rely on a single fixture or plasma-only fallback. \\
015 & Biomedical Engineering & \href{https://github.com/IlyaKuprov/Spinach}{\nolinkurl{IlyaKuprov/Spinach}} & Issue \#22 & spin-dynamics thermalization, formalism-specific identity/projector construction, unit-state representation consistency & thermalize should use formalism-specific identity/projector, and unit\_state cannot be treated as two objects at the same time. \\
016 & Biology & \href{https://github.com/macs3-project/MACS}{\nolinkurl{macs3-project/MACS}} & Issue \#735 / PR \#736 & HMMRATAC semantics, blacklist filtering, query EOF post-count consistency & HMMRATAC refinement must first guard empty peak\_content, and cannot close empty peaks on sparse bedGraph. \\
017 & Astronomy & \href{https://github.com/gammapy/gammapy}{\nolinkurl{gammapy/gammapy}} & Issue \#6754 / PR \#6757 & J-factor line-of-sight impact parameter, b = D sin(theta), r = b cosh(t) stable integration & The J-factor line integral must stabilize the branch according to the impact parameter, and cannot use tan(theta) to push directly to the singular point. \\
018 & Biology & \href{https://github.com/macs3-project/MACS}{\nolinkurl{macs3-project/MACS}} & Issue \#680 / PR \#689 & blacklist filtering, HMMRATAC training/inference path, peak call statistical consistency & The blacklist should be filtered before training and segmentation, and blacklist segments cannot be allowed to enter the model and counting of HMMRATAC. \\
019 & Biology & \href{https://github.com/aertslab/pycisTopic}{\nolinkurl{aertslab/pycisTopic}} & PR \#196 & Discretization of topic probability matrix to region / cell set, legacy object API, Regions\_in\_binarized\_topic maintenance & Topic binarization needs to support both matrix/object entries and stable threshold semantics, and cannot just cover the old object API. \\
020 & Materials Science and Engineering & \href{https://github.com/deepmodeling/deepmd-kit}{\nolinkurl{deepmodeling/deepmd-kit}} & PR \#5581 & neighbor graph, periodic-cell mapping, segment reductions, force / atom virial scatter & The neighbor graph needs to unify PBC mapping, empty segment reduction and force/virial scatter, and guard edges cannot be regarded as effective contributions. \\
021 & Chemistry & \href{https://github.com/deepmodeling/dpti}{\nolinkurl{deepmodeling/dpti}} & PR \#103 & coupling endpoints, pair-specific epsilon, three-stage HTI/FEP input generation, thermo\_style consumption & Soft-core LJ FEP must generate a real \texttt{compute fep}; energy divided by lambda cannot substitute for derivative observations. \\
022 & Biomedical Engineering & \href{https://github.com/nilearn/nilearn}{\nolinkurl{nilearn/nilearn}} & Issue \#5168 / PR \#5169 & Volume-to-surface atlas projection, discrete region identity preservation under nearest sampling, missing samples should return NaN & Discrete brain area projections should adopt mode sampling, and NaN should be retained for missing samples. Labels cannot be averaged numerically. \\
023 & Chemistry & \href{https://github.com/MDAnalysis/mdanalysis}{\nolinkurl{MDAnalysis/mdanalysis}} & Issue \#4257 / PR \#4258 & NPT trajectory without jump expansion, box dimensions update, first frame displacement history initialization & NoJump's frame bookkeeping must preserve the first cross-boundary displacement and cannot exit early at frame 0. \\
024 & Statistics & \href{https://github.com/pymc-devs/pymc}{\nolinkurl{pymc-devs/pymc}} & Issue \#6871 / PR \#6872 & ZeroSumNormal's log-density, normalization and pseudo-determinants on zero-sum manifolds & The logp of ZeroSumNormal must be calculated on the zero-sum manifold and cannot directly reuse ambient Gaussian normalization. \\
025 & Biomedical Engineering & \href{https://github.com/schorschinho/osprey}{\nolinkurl{schorschinho/osprey}} & Issue \#447 / PR \#448; Issue \#657 / PR \#658 & acquisition metadata consistency in the edited-MRS workflow, consistent calibration reports and representations & Edited-MRS acquisition metadata and the water-calibration time scale must remain consistent; TE/TR must not be converted a second time. \\
026 & Atmospheric Science & \href{https://github.com/Unidata/MetPy}{\nolinkurl{Unidata/MetPy}} & Issue \#3315 / PR \#3324 & Isentropic interpolation on sigma coordinates, pressure column selection, dimension semantic correction & Isentropic interpolation must respect vertical coordinate semantics such as sigma/eta, and cannot treat them directly as pressure columns. \\
027 & Chemistry & \href{https://github.com/little1d/NMRTrans}{\nolinkurl{little1d/NMRTrans}} & Issue \#4 & Masking and representation semantics of NMR peak-set encoding, batch padding, and set transformer & The mask semantics of the peak encoder are reversed, the padding is retained, the real peaks are masked, and the set representation is distorted. \\
028 & Biology & \href{https://github.com/JialongJiang/DSPIN}{\nolinkurl{JialongJiang/DSPIN}} & Not specified & control-referenced relative responses, perturbation response normalization and fallback path integrity & The control sample index and relative response baseline cannot be misaligned, otherwise the disturbance response will shift as a whole. \\
029 & Biology & \href{https://github.com/biopython/biopython}{\nolinkurl{biopython/biopython}} & Issue \#4611 / PR \#4637 & Feature coordinates and sequence semantics after circular record orientation normalization and reverse-complement & After circular feature reverse-complement, the part order must be reversed, and the original cross-origin order cannot be retained. \\
030 & Materials Science and Engineering & \href{https://github.com/materialsproject/pymatgen}{\nolinkurl{materialsproject/pymatgen}} & Issue \#3677 / PR \#3779 & SHELX/AIRSS.res export, SFAC ordered species column, stable round-trip serialization & The SFAC of.res is an ordered list, and species cannot be written as an unordered set. \\
031 & Biomedical Engineering & \href{https://github.com/mne-tools/mne-python}{\nolinkurl{mne-tools/mne-python}} & Issue \#13913 / PR \#13920 & average reference across multiple sensor types, classification type re-reference and joint / projection path consistency & The average reference of mixed sensor types should be done separately according to channel type and cannot be directly combined into a union mean. \\
032 & Chemistry & \href{https://github.com/openmm/openmm}{\nolinkurl{openmm/openmm}} & Issue \#5025 / PR \#5026 & GROMACS topology import, NBFIX / combination rule semantics, total energy representation maintenance & GROMACS topology import needs to unify the translation semantics of NBFIX, combination rule and 1-4 exception. \\
033 & Materials Science and Engineering & \href{https://github.com/pyscf/pyscf}{\nolinkurl{pyscf/pyscf}} & PR \#2946 & consistency between k-point and Gamma-supercell representations in periodic HF, Coulomb wrap-around semantics & Periodic HF k-point and Gamma-supercell paths must implement equivalent Coulomb semantics; wrap-around handling must not diverge between them. \\
034 & Mathematics & \href{https://github.com/patrick-kidger/diffrax}{\nolinkurl{patrick-kidger/diffrax}} & Issue \#598 / PR \#746 & strong order, reverse-oriented SDE integration, boundary conditions and hidden state processing & The step size and Levy area parity of the reverse time SDE must be processed according to the direction, and the forward SRK convention cannot be followed. \\
035 & Mathematics & \href{https://github.com/scipy/scipy}{\nolinkurl{scipy/scipy}} & Issue \#14480 / PR \#14552 & Numerical consistency of LSODA event processing, dense output, stiff kinetics & The dense output of LSODA needs to take the history of the last successful step, and cannot use the next candidate state for event positioning. \\
036 & Marine Science & \href{https://github.com/Parcels-code/Parcels}{\nolinkurl{Parcels-code/Parcels}} & Issue \#2521 / PR \#2598 & Problems with spherical curvilinear grids, latitude and longitude grid index, and downstream parameters still follow the old path & Curved grid point-in-cell must be calculated according to spherical geometry and cannot continue to be approximated by longitude and latitude planes. \\
037 & Biomedical Engineering & \href{https://github.com/nipy/nibabel}{\nolinkurl{nipy/nibabel}} & PR \#1340 & Enhanced multiframe DICOM reconstruction, frame/direction semantics and image plane consistency & The frame order of multi-frame DICOM must be consistent with the physical slice order, and cannot be reconstructed solely by index. \\
038 & Materials Science and Engineering & \href{https://github.com/spglib/spglib}{\nolinkurl{spglib/spglib}} & Issue \#584 / PR \#589 & Classification stability, primitive/conventional transformation and symmetry search under equivalent lattice representation & The operation route of symmetry search must be updated synchronously with the base transformation and cannot be disconnected from the structure route. \\
039 & Geography & \href{https://github.com/landlab/landlab}{\nolinkurl{landlab/landlab}} & Issue \#2408 / PR \#2409 & parcel transport, reach-scale flux consistency, segmentation and external representation consistency & The speed of parcel transport must be consistent with reach-level flux conversion, and the number of particles cannot be allowed to change the physical transport rate. \\
040 & Geophysics & \href{https://github.com/simpeg/simpeg}{\nolinkurl{simpeg/simpeg}} & Issue \#1099 / PR \#1575 & magnetic-dipole forward modeling, source representation equivalence, numerical accuracy and theoretical fit & Different formulations of magnetic dipole forward modeling must fall into the same discrete field space, and the representation method cannot change the results. \\
041 & Geophysics & \href{https://github.com/sunpy/sunpy}{\nolinkurl{sunpy/sunpy}} & Issue \#8406 / PR \#8415 & Earth-centered coordinate transformations, GEI/GSE path consistency and vector conservation & The GEI/GSE transformation must follow a purely rotational path and cannot go through an intermediate map with additional corrections. \\
042 & Astronomy & \href{https://github.com/sunpy/sunpy}{\nolinkurl{sunpy/sunpy}} & Issue \#8490 / PR \#8548 & Sun coordinate objects maintain frame, observer, obstime, representation, units and numerical meanings in table serialization & The solar coordinate table round-trip must retain frame, observer and obstime, and cannot compress the coordinate object into a pure numerical value. \\
043 & Materials Science and Engineering & \href{https://github.com/PMEAL/porespy}{\nolinkurl{PMEAL/porespy}} & Issue \#966 / PR \#1174 & Two-point correlation function, aperiodic boundaries, overlap counting and radial binning on finite images & Finite image two-point correlation needs to perform overlap and radial bin statistics according to finite-image semantics, and cannot be treated as a periodic domain. \\
044 & Chemistry & \href{https://github.com/pybamm-team/PyBaMM}{\nolinkurl{pybamm-team/PyBaMM}} & PR \#5600 & In regularized power operations, distinguish between state-independent fitting coefficients and dynamic state-dependent expressions & The fractional-power stabilization in the exchange current expression must distinguish between constants and state variables, and the parameters themselves cannot be rewritten. \\
045 & Marine Science & \href{https://github.com/xgcm/xgcm}{\nolinkurl{xgcm/xgcm}} & Issue \#712 / PR \#713 & Connection surface direction, axis order, consistency of pad and transpose & cubed-sphere halo/padding must first align the dimensions and surface directions of the source slice, and cannot read the adjacent surfaces incorrectly after transpose. \\
046 & Astronomy & \href{https://github.com/StingraySoftware/stingray}{\nolinkurl{StingraySoftware/stingray}} & PR \#965 & Coherence/power spectrum calculation under effective domain, boundary state and invalid value determination & The coherence estimate must correctly deduct the noise deviation and count as averaged spectra, and cannot return a finite value in the noise-dominated frequency band. \\
047 & Information and Communication Engineering & \href{https://github.com/scikit-rf/scikit-rf}{\nolinkurl{scikit-rf/scikit-rf}} & Issue \#654 / PR \#655 & Conversion and parameter consistency of different microwave network representations under complex reference impedance & Network connections under complex reference impedance must first undergo wave-definition transformation, and the traveling-wave convention cannot be used. \\
048 & Chemistry & \href{https://github.com/pycalphad/pycalphad}{\nolinkurl{pycalphad/pycalphad}} & Issue \#310 / PR \#311 & consistency between phase energy, Gibbs energy, and ordered-phase representation & Gibbs-energy bookkeeping for ordered/disordered phases must project onto the non-ordering sublattice; database encoding must not change phase stability. \\
049 & Mathematics & \href{https://github.com/zwicker-group/py-pde}{\nolinkurl{zwicker-group/py-pde}} & PR \#839 & Multi-step physical process semantics and incombinability in the evolution of discrete random fields & The noise terms and volume factors of discrete random fields must be processed hierarchically, and cell-volume scaling cannot be placed in the wrong location. \\
050 & Physics & \href{https://github.com/jcmgray/quimb}{\nolinkurl{jcmgray/quimb}} & Issue \#402 & Scale-dependent norm reconstruction, result consistency of different tensor representations & The norm of the tensor network must remain unchanged after gauging/rescaling, and the approximate contraction cannot change the exact norm. \\
051 & Geophysics & \href{https://github.com/SHTOOLS/SHTOOLS}{\nolinkurl{SHTOOLS/SHTOOLS}} & PR \#397 & Spherical harmonic gradient, polar boundary and planetary magnetic field calculation & The gravity/magnetic force second-order tensor must be consistent from the local frame to the Cartesian frame, and recursion and boundary constraints cannot be allowed to destroy the rotation. \\
052 & Materials Science and Engineering & \href{https://github.com/usnistgov/fipy}{\nolinkurl{usnistgov/fipy}} & Issue \#1190 / PR \#1192 & Modular gradients and physical geometry consistency on non-uniform orthogonal grids & Non-uniform Grid3D needs to unify cell widths, face sizes and modular gradient semantics, and cannot only retain a single spacing. \\
053 & Biomedical Engineering & \href{https://github.com/Project-MONAI/MONAI}{\nolinkurl{Project-MONAI/MONAI}} & Issue \#4103 / PR \#6681 & Physical spacing, subvoxel surface area approximation and surface Dice & Surface Dice should be calculated based on physical surface area, and anisotropic voxels cannot simply be regarded as equal-area boundaries. \\
054 & Chemistry & \href{https://github.com/matchms/matchms}{\nolinkurl{matchms/matchms}} & PR \#863 & Screening, merging and entropy-based spectral similarity of highly similar peaks & Entropy similarity cannot allow overlapping candidate windows to count the same peak repeatedly, and the pair/matrix paths must share the same peak semantics. \\
055 & Astronomy & \href{https://github.com/astropy/specutils}{\nolinkurl{astropy/specutils}} & Issues \#817 / \#1059 / \#1300; PRs \#1060 / \#1305 & Boundary extension, partially overlapping bins and finite interval filling & Spectral resampling must propagate flux and variance according to limited bin overlap, and bins cannot be used as point samples. \\
056 & Aeronautical and Astronautical Science and Technology & \href{https://github.com/poliastro/poliastro}{\nolinkurl{poliastro/poliastro}} & Issue \#1563; PR \#1583 & Consistency between orbital state and binding epoch, history sampling semantics & Each state and epoch of Orbit->Ephem must point to the same orbital state, and the anomaly cannot drift with the timeline. \\
057 & Chemistry & \href{https://github.com/mdtraj/mdtraj}{\nolinkurl{mdtraj/mdtraj}} & Issue \#1925; PR \#1929 & Molecular reconstruction under periodic boundaries, bonded geometry and coordinate order independence & make\_molecules\_whole restores the periodic image along the bond graph in one pass, and cannot let the bond order determine the translation results. \\
058 & Nuclear Science and Technology & \href{https://github.com/openmc-dev/openmc}{\nolinkurl{openmc-dev/openmc}} & Issue \#3852; PR \#3853 & Rectangular lattice distance, boundary and geometric determination in particle transport & The boundary distance and transport event of the repeated rectangular lattice must be calculated correctly, and the legal geometry cannot be misjudged as terminated. \\
059 & Computer Science and Technology & \href{https://github.com/scikit-image/scikit-image}{\nolinkurl{scikit-image/scikit-image}} & Issue \#5112; PR \#6296 & voxel spacing, area and morphometric calibration of regionprops & regionprops should treat spacing as a physical scale and cannot treat anisotropic sampling as a lattice index. \\
060 & Biomedical Engineering & \href{https://github.com/SpikeInterface/spikeinterface}{\nolinkurl{SpikeInterface/spikeinterface}} & Issue \#3328; PR \#3517 & non-equidistant sampling, cross-band motion correction, and interpolation correction & Motion interpolation must align AP/LF streams on the physical time axis; assuming uniform motion from the sampling rate is invalid. \\
061 & Astronomy & \href{https://github.com/astropy/reproject}{\nolinkurl{astropy/reproject}} & Issue \#199; PR \#589 & Numerical elimination, reprojection and energy conservation of small sphere area & The extremely small sky pixel in high-resolution reproject will have numerical cancellation, and overlap geometry and flux conservation must be stable. \\
062 & Materials Science and Engineering & \href{https://github.com/materialsproject/atomate2}{\nolinkurl{materialsproject/atomate2}} & Issue \#333 / PR \#415 & Green-Lagrange strain-stress fitting, independent shear response reconstruction, and tensor symmetry in a high-throughput crystal elasticity workflow & Elastic tensor reconstruction must bring the complete shear response back to the final result, and the reduce/reconstruct link cannot lose symmetry information. \\
063 & Biology & \href{https://github.com/tskit-dev/tskit}{\nolinkurl{tskit-dev/tskit}} & Issue \#3053, \#3175 / PR \#3059, \#3176 & Distinguishing between "no lineage" and "multi-rooted trees with no common ancestors" under window normalization, and normalizing the observation span & Window normalized pair-coalescence needs to distinguish between non-lineage trees and multi-rooted trees without common ancestors, and the boundary semantics of the two types cannot be confused. \\
064 & Chemistry & \href{https://github.com/cclib/cclib}{\nolinkurl{cclib/cclib}} & Issue \#660 / PR \#745 & Signed energy extraction, unit consistency and ccData exposure for DFT-D dispersion correction in ORCA output & ORCA's DFT-D dispersion correction should be written out as an independent signed physical quantity and not just left inside the parser. \\
065 & Astronomy & \href{https://github.com/yt-project/yt}{\nolinkurl{yt-project/yt}} & Issue \#4788 / PR \#4939 & SPH-based three-dimensional projection, oblique slicing, real sight integral and kernel function geometric semantics & SPH line-of-sight integration and oblique slicing need to share consistent kernel geometry and normalization, and the resolution cannot be allowed to change the observation results. \\
066 & Atmospheric Science & \href{https://github.com/SciTools/iris}{\nolinkurl{SciTools/iris}} & Issue \#4830 / PR \#5548 & Equivalent identification of Rotated Mercator and Oblique Mercator under CF grid mapping, projection object and site matching semantics & The CF grid mapping of rotated Mercator and oblique Mercator should be recognized as equivalent and cannot be linked to the wrong projection reference. \\
067 & Atmospheric Science & \href{https://github.com/Ouranosinc/xclim}{\nolinkurl{Ouranosinc/xclim}} & Issue \#700 / PR \#704 & Overwintering precipitation and Drought Code state transfer, recursive boundary between active season and dormant phase when seasons switch & The seasonal state machine of the fire danger index should separate the winter precipitation and active-day updates, and the state transfer cannot be connected to the wrong stage. \\
068 & Biology & \href{https://github.com/samtools/htslib}{\nolinkurl{samtools/htslib}} & Issue \#1820 / PR \#1897 & Actual coverage interval, rlen, END / symbolic allele / reference block semantics recorded on the reference genome & The reference occupancy interval of VCF/BCF should be consistent on paths such as symbolic allele, END and SVLEN. \\
069 & Biology & \href{https://github.com/open2c/cooltools}{\nolinkurl{open2c/cooltools}} & Issue \#456 / PR \#501 & Contact-frequency and smoothed P(s) propagation, normalization and multi-region aggregation consistency & expected cis smoothing needs to propagate the numerators and denominators of raw contact and weighted contact separately, and observation apertures cannot be mixed. \\
070 & Biology & \href{https://github.com/levitsky/pyteomics}{\nolinkurl{levitsky/pyteomics}} & Issue \#202 / PR \#203 & ProForma local modification and global ion-carrying charge calculation, charge\_state consistency & ProForma's global loads, local modifications, and neutral masses/m/z require unified charge accounting, and electrons cannot be counted repeatedly. \\
071 & Physics & \href{https://github.com/silx-kit/pyFAI}{\nolinkurl{silx-kit/pyFAI}} & Issue \#2773 / PR \#2772 & Transformation from detector direction to sample coordinate system q vector, exit-angle and reciprocal represent consistency & The q vector transformation must maintain the consistency of the detector to sample coordinates, and the orientation encoding cannot change the reciprocal space results. \\
072 & Biomedical Engineering & \href{https://github.com/pydicom/pydicom}{\nolinkurl{pydicom/pydicom}} & Issue \#2210 / PR \#2212 & Bit depth, sampling, limited-range luma/chroma conversion of YBR\_FULL / YBR\_FULL\_422 / YBR\_PARTIAL\_420 & The YBR\_* pixel format must unify bit depth and sampling extension semantics, and cannot just modify the limited-range path. \\
073 & Biology & \href{https://github.com/biotite-dev/biotite}{\nolinkurl{biotite-dev/biotite}} & Issue \#777 / PR \#781 & Conformational connectivity, bond map preservation and bond mapping after alternate location selection & The alternate location filtering must be completed before the bond graph, and the selected conformation cannot be disconnected from the struct\_conn. \\
074 & Materials Science and Engineering & \href{https://github.com/deepmodeling/DeePTB}{\nolinkurl{deepmodeling/DeePTB}} & Issue \#332 / PR \#320 & Generalized eigenvalue problems, ill-conditioned overlap matrix processing, finite band data and validity masks & The generalized eigenvalue problem of non-orthogonal tight binding must first deal with the ill-posed overlap matrix, and cannot directly Cholesky on the invalid subspace. \\
075 & Materials Science and Engineering & \href{https://github.com/deepmodeling/deepmd-kit}{\nolinkurl{deepmodeling/deepmd-kit}} & Issue \#5593 / PR \#5601 & Boundary interpolation, first-order sensitivity, layout and derivative consistency of DeePMD compressed tables & Compressed table extrapolation must maintain both value and derivative continuation semantics, and C1 continuation cannot be truncated outside the boundary. \\
076 & Mathematics & \href{https://github.com/cvxpy/cvxpy}{\nolinkurl{cvxpy/cvxpy}} & Issue \#3092 / PR \#3093 & Coordinate mapping of N-dimensional parameters to sparse coefficient matrix after indexing, broadcasting, and matrix multiplication & Canonicalization of the Parameter tensor requires first folding repeated positions and then mapping to a sparse coefficient matrix. \\
077 & Mathematics & \href{https://github.com/shapely/shapely}{\nolinkurl{shapely/shapely}} & Issue \#1670, \#1929 / PR \#1708, \#1939 & Numerical stability fallback, candidate edge enumeration, NumPy ufunc batching and output buffering compatibility & Oriented envelope must truly select the smallest area among candidate edges and is compatible with NumPy batch processing. \\
078 & Astronomy & \href{https://github.com/astropy/photutils}{\nolinkurl{astropy/photutils}} & Issue \#2287 / PR \#2288 & Bidirectional transformation, direction consistency and round-trip constraints of aperture geometry under WCS & The conversion from sky aperture to pixel aperture must keep the WCS local base direction consistent, and the aperture mirror cannot be shifted. \\
079 & Chemistry & \href{https://github.com/aspuru-guzik-group/selfies}{\nolinkurl{aspuru-guzik-group/selfies}} & Issue \#120 / PR \#125 & kekulization of SMILES aromatic system to SELFIES, charged aromatic carbon electron counting & When converting the SMILES aromatic system to SELFIES, the electron count of the charged aromatic carbon must also be dealt with, and the neutral zero-hydrogen path cannot only be repaired. \\
080 & Surveying and Mapping Science and Technology & \href{https://github.com/xarray-contrib/xarray-spatial}{\nolinkurl{xarray-contrib/xarray-spatial}} & Issue \#1086 / PR \#1087 & footprint, coordinate direction normalization, nodata propagation and cubic boundary interpolation & The footprint, coordinate direction and nodata propagation during reprojection must be bound to the interpolation order, and the boundaries cannot be allowed to degenerate into copied values. \\
081 & Biology & \href{https://github.com/scverse/anndata}{\nolinkurl{scverse/anndata}} & Issue \#1854 / PR \#2122 & backed HDF5/Zarr re-indexing, outer join, and sparse/dense annotation consistency & Disk-backed concatenation must write non-concatenated-axis annotations back under the new index; preserving only the main matrix is insufficient. \\
082 & Materials Science and Engineering & \href{https://github.com/deepmodeling/dpgen}{\nolinkurl{deepmodeling/dpgen}} & Issue \#1639 / PR \#1801 & DP-GEN active-learning input generation, LAMMPS template revise, model-deviation control items, relative error and electron temperature propagation & DP-GEN must propagate the same set of LAMMPS model-deviation settings in both direct generation and template revision paths, and the relative, cluster or electron temperature items cannot be lost in the template path. \\
083 & Chemistry & \href{https://github.com/deepmodeling/reacnetgenerator}{\nolinkurl{deepmodeling/reacnetgenerator}} & Issue \#2165 / PR \#2221 & bond-order trajectory, atom-bond graph equivalence, isomer merge mode, cyclic structure species naming and reaction network report & The species report of ReacNetGenerator must define the same molecule according to chemical graph equivalence and merge mode, and cannot separate bond order changes or traversal order into different species. \\
084 & Materials Science and Engineering & \href{https://github.com/deepmodeling/dpdata}{\nolinkurl{deepmodeling/dpdata}} & Issue \#951 / PR \#957 & ABACUS relaxation-output parsing, frame-wise coordinates and cell labels, archive/reload round-trip, complete trajectory recovery & The ABACUS relaxation round-trip must preserve the complete relaxation history and frame-wise labels, rather than retaining only the final structure or losing cell/coordinate semantics. \\
085 & Chemistry & \href{https://github.com/dftbplus/dftbplus}{\nolinkurl{dftbplus/dftbplus}} & Issue \#1069 / PR \#1462 & DFTB+ COSMO/GB archived observations, atomic order invariance, equivalent atomic records and solvent surface adjacencies & DFTB+ implicit solvent observations should be insensitive to equivalent atoms and input order, and should not allow the order of atom records to change COSMO-like reporting. \\
086 & Materials Science and Engineering & \href{https://github.com/materialyzeai/matgl}{\nolinkurl{materialyzeai/matgl}} & Issue \#642 / PR \#819 & MatGL dataset element list, pre-trained model element vocabulary, graph construction atom type encoding, fine-tuning compatibility check & MatGL fine-tuning must first verify that the data set element list is consistent with the model element vocabulary, and the graph construction cannot be allowed to silently map atoms to wrong embeddings. \\
087 & Biomedical Engineering & \href{https://github.com/dipy/dipy}{\nolinkurl{dipy/dipy}} & Issue \#3968 / PR \#3976 & Patch2Self small batch denoising, trailing-volume flush, b0 volume data processing, patch radius and baseline shift & Patch2Self wants to keep the diffuse volume consistent with b0 semantics on mini-batch and tail volume data, and cannot leave finite but offset denoised observations. \\
088 & Atmospheric Science & \href{https://github.com/pydata/xarray}{\nolinkurl{pydata/xarray}} & Issue \#9108 / PR \#9116 & CFTimeIndex resample, calendar and has\_year\_zero reserved, year-zero near month/year-end boundary aggregation & CF time aggregation must retain calendar and has\_year\_zero semantics, and cannot aggregate legal short sequences into failure paths at the year-zero boundary. \\
089 & Chemistry & \href{https://github.com/cclib/cclib}{\nolinkurl{cclib/cclib}} & Issue \#46 / PR \#1085 & ORCA excited-state absorption spectrum, oscillator strength, multiple transition attribute table, etoscs representation selection & cclib must retain the correct oscillator-strength representation when parsing ORCA excited state spectra, and must not allow subsequent or substitute spectra to overwrite semantic values. \\
090 & Chemistry & \href{https://github.com/jjhelmus/nmrglue}{\nolinkurl{jjhelmus/nmrglue}} & Issue \#187 / PR \#189 & Bruker NUS compressed FID stream, nuslist sparse coordinates, quadrature/acquisition order, missing FID zero-fill & nmrglue needs to expand the sparse FID stream into a multi-dimensional NMR grid according to the nuslist and collection order, and cannot treat the compressed stream as an ordinary dense 2D sequence. \\
091 & Materials Science and Engineering & \href{https://github.com/materialsproject/pymatgen-core}{\nolinkurl{materialsproject/pymatgen-core}} & Issue \#4607 / PR \#87 & Gaussian cube nonzero origin, Bohr-to-angstrom unit conversion, voxel origin and structural coordinate placement & When reading Gaussian cube, the non-zero volume data origin and unit conversion must be included in the structure placement. The first voxel cannot be the default coordinate origin. \\
092 & Chemistry & \href{https://github.com/chemprop/chemprop}{\nolinkurl{chemprop/chemprop}} & Issue \#426 / PR \#427 & Condensed Graph of Reaction, atom mapping, reaction-mode options, and reactant/product feature propagation & Chemprop reaction featurization must construct a CGR according to the reaction mode and atom map; mapped atoms must be retained and reactant/product changes must not be encoded in reverse. \\
093 & Biomedical Engineering & \href{https://github.com/Project-MONAI/MONAI}{\nolinkurl{Project-MONAI/MONAI}} & Issue \#6985 / PR \#7000 & ITKReader, MetaTensor, ITK image round-trip, affine/spacing/origin/direction coordinate convention & MONAI's reader -> MetaTensor -> ITK round-trip must preserve patient-space coordinates and orientation, image geometry cannot be silently changed between equivalent paths. \\
094 & Physics & \href{https://github.com/materialsproject/pymatgen}{\nolinkurl{materialsproject/pymatgen}} & PR \#1706 & magnetic space group operations, BNS/Jones-Faithful setting transform, fractional translation, time reversal & Magnetic space group archive records must be converted according to setting to preserve translation and time-reversal semantics, and the old coordinate convention cannot be read directly as the current convention. \\
095 & Materials Science and Engineering & \href{https://github.com/PMEAL/OpenPNM}{\nolinkurl{PMEAL/OpenPNM}} & Issue \#2843 / PR \#2849 & throat conductance field, model builder, operator assembly, advection-diffusion solver, repeated stage transfer & Porous network transmission requires consistent transmission of conductance representation from model output to solver and flux calculations. Object fields and array fields cannot be mixed in repeated solves. \\
096 & Atmospheric Science & \href{https://github.com/CLIMADA-project/climada_python}{\nolinkurl{CLIMADA-project/climada_python}} & Issue \#831 / PR \#839 & antimeridian longitude normalization, track bounds, centroid filtering, sparse hazard matrix, equivalent date-line encoding & Tropical cyclone hazard calculations need to unify longitude wrapper semantics in paths, centroid ranges, and sparse maps, and cannot miss or misplace wind fields near \ensuremath{\pm}180 degrees. \\
097 & Physics & \href{https://github.com/deepmodeling/deepmd-kit}{\nolinkurl{deepmodeling/deepmd-kit}} & Issue \#5663 / PR \#5854 & Spin virtual types, negative placeholder atoms, dense/lower layouts, dpmodel/PT parity, padded mixed-size batches & DeePMD spin model preprocessing requires negative type padding to remain inert in real/virtual type expansion and backend layout, and cannot change placeholder atoms into valid spin types. \\
098 & Surveying and Mapping Science and Technology & \href{https://github.com/pytroll/satpy}{\nolinkurl{pytroll/satpy}} & Issue \#3182 / PR \#3217 & geostationary satellite position units, TLE fallback, cloud-top-height parallax, observer-look boundary, inverse resampling & Satpy parallax correction needs to unify internal satellite height/radius units, only convert at external library boundaries, and cannot mix meters/km between reader metadata, TLE and Cartesian geometry. \\
099 & Surveying and Mapping Science and Technology & \href{https://github.com/Stanford-NavLab/gnss_lib_py}{\nolinkurl{Stanford-NavLab/gnss_lib_py}} & PR \#121 & WLS pseudorange positioning, Earth-rotation/Sagnac correction, satellite ECEF convention, simulator direct satellite states & GNSS WLS point positioning requires rotating the satellite position to the receive measurement frame and being compatible with simulator input. Transmitted-frame ECEF cannot be directly solved as receive-frame. \\
100 & Chemistry & \href{https://github.com/DCoupry/autografs}{\nolinkurl{DCoupry/autografs}} & Issue \#174/\#197/\#198/\#207 / PR \#208 & AuToGraFS relaxed embedding, topology slot motion, closure regression, cell volume scaling, pinned-net control & Relaxed framework embedding must allow low symmetry slot movement while maintaining topology, closure and volume scale, the objective cannot be adjusted to allow individual nets to pass. \\
101 & Chemistry & \href{https://github.com/pybamm-team/PyBaMM}{\nolinkurl{pybamm-team/PyBaMM}} & Issue \#5414 / PR \#5540 & half-cell Doyle-Fuller-Newman model, surface-form electrolyte potential, boundary-condition registration, thermal Ohmic-heating coupling & PyBaMM half-cell DFN must consistently register electrolyte/solid phase potential boundary conditions for surface-form and thermal sub-models, and cannot allow half-cell domain splitting to cause gradient shape mismatch. \\
102 & Materials Science and Engineering & \href{https://github.com/marrink-lab/polyply_1.0}{\nolinkurl{marrink-lab/polyply_1.0}} & Issue \#380 / PR \#381 & polymer residue graph, template/volume identity, coordinate generation, library loading, residue equivalence & Polymer workflows must pass template, volume, and residue graph identities all the way to coordinate generation without losing graph structure semantics between loading and generation. \\
103 & Mechanics & \href{https://github.com/SteveDoyle2/pyNastran}{\nolinkurl{SteveDoyle2/pyNastran}} & Issue \#584 / PR \#85 & symmetric PCOMP expansion, ABD equivalent, D block calculation, explicit layup and symmetric layup consistency & The ABD of the symmetrical layup must be consistent with the explicit layup, and the expanded directional stiffness and D matrix cannot deviate. \\
104 & Chemistry & \href{https://github.com/pybamm-team/PyBaMM}{\nolinkurl{pybamm-team/PyBaMM}} & Issue \#3526 / PR \#3979 & upwind/downwind edge states, cell-centered and edge-centered flux, Dirichlet FV boundary handling, mesh-size invariance & The three-layer semantics of cell/edge/boundary in a finite volume must be consistent, and upwind/downwind cannot be mismatched in boundary processing. \\
105 & Chemistry & \href{https://github.com/rdkit/rdkit}{\nolinkurl{rdkit/rdkit}} & Issue \#9338 / PR \#9339 & tautomer canonicalization, stereo preservation, V3000 structure serialization, atropisomer bond validity & Tautomeric normalization must preserve chirality and output serializable molecules, and the stereochemistry cannot be broken in canonicalize. \\
106 & Chemistry & \href{https://github.com/MLCIL/scikit-fingerprints}{\nolinkurl{MLCIL/scikit-fingerprints}} & Issue \#519 / PR \#525 / PR \#528 & MAP4 / MAPC retrieval protocol, chirality invariance, representation consistency, benchmark panel, version pinning & Molecular retrieval protocols need to maintain the same semantics across chirality, representation, and batch processing paths, and cannot be true only for visible panels. \\
107 & Mathematics & \href{https://github.com/sympy/sympy}{\nolinkurl{sympy/sympy}} & Issue \#28186 / PR \#28189 & rational integration, exact root handling, log\_to\_real, derivative identity, real antiderivative semantics & The original function returned by rational integration must really be able to convert back to the original integrand, and root processing cannot break the derivative identity. \\
108 & Mathematics & \href{https://github.com/mpmath/mpmath}{\nolinkurl{mpmath/mpmath}} & Issue \#1104 / PR \#1107; Issue \#930 / PR \#1111 & theta3/theta4 fixed-precision paths, near-unit nome stability, tiny-nome cancellation, derivative propagation, modular transform & Jacobi theta must also be stable on legal nomes. It cannot be rejected by mistake, nor can it explode on the path of small nomes. \\
109 & Mechanics & \href{https://github.com/elhajjar1/wrinkleFE}{\nolinkurl{elhajjar1/wrinkleFE}} & Issue \#252 / PR \#285 & multi-wrinkle field composition, fiber-angle propagation, phase offsets, FE analysis, RSS shortcut & The multi-wrinkle field cannot be approximated by single wrinkle RSS, and the fiber angle must follow the synthesized wrinkle field. \\
110 & Civil Engineering & \href{https://github.com/robbievanleeuwen/section-properties}{\nolinkurl{robbievanleeuwen/section-properties}} & Issue \#272 / PR \#287 & first-moment result, outline quantity semantics, cross-section property consistency, reported vs constructed geometry & The first-order moment result cannot be mistakenly reported as an outline quantity, and the report semantics and construction geometry must be consistent. \\
111 & Mechanics & \href{https://github.com/GazzolaLab/PyElastica}{\nolinkurl{GazzolaLab/PyElastica}} & Issue \#81 / PR \#87 & tapered slender filaments, inverse-time material-response, cross-section variation, translational observations, free-response consistency & The free response of the slender body cannot jump randomly with the change of cross section, and the material response and geometric changes must be coordinated. \\
112 & Civil Engineering & \href{https://github.com/robbievanleeuwen/section-properties}{\nolinkurl{robbievanleeuwen/section-properties}} & Issue \#460 / PR \#462 & hybrid-section plastic capacity, constituent force/centroid preservation, split geometry pieces, plastic neutral axis & The plastic load-bearing of the composite section must preserve the force and centroid of each material segment, and cannot just report a seemingly reasonable number. \\
113 & Electrical Engineering & \href{https://github.com/pvlib/pvlib-python}{\nolinkurl{pvlib/pvlib-python}} & Issue \#2054 / PR \#2055 & timezone-aware timestamps, calendar bins, day-of-year, extraterrestrial irradiance, SPA, equivalent aware representations & Solar resource calculations must give consistent results for "different time representations of the same moment", and time zone normalization cannot be aliased. \\
114 & Electrical Engineering & \href{https://github.com/PyPSA/PyPSA}{\nolinkurl{PyPSA/PyPSA}} & Issue \#1360 / PR \#1371 & storage history, planning-epoch boundary, cyclic vs initial state, Store/StorageUnit boundary semantics, state recurrence & The energy storage state must abide by the cycle boundary priority rules before and after the planning boundary, and the initial state cannot be mistaken for the previous period. \\
115 & Civil Engineering & \href{https://github.com/JWock82/Pynite}{\nolinkurl{JWock82/Pynite}} & Issue \#255, \#275 / PR \#323 & Global stiffness singularity, structural stability determination, sparse/dense solution consistency, zero right end boundary & Legal frames cannot return finite but distorted displacements when unstable. Global stiffness singularities, zero load boundaries, and sparse/dense path consistency must be determined together. \\
116 & Civil Engineering & \href{https://github.com/JWock82/Pynite}{\nolinkurl{JWock82/Pynite}} & Issue \#255, \#275 / PR \#323 & Structural response reports, equilibrium equations, matrix storage modes, physical interpretation of control studies & Response reporting by the same framework in sparse / dense matrix storage modes must remain physically interpretable and consistent with equilibrium equations and API conventions. \\
117 & Civil Engineering & \href{https://github.com/JWock82/Pynite}{\nolinkurl{JWock82/Pynite}} & Issue \#317 / PR \#319 & inactive member results, node displacement chord, physical member segmentation, local/global coordinate transformation & The inactive member still needs to be linearly interpolated along the chord after displacement to ensure that the scalar / sampled results are compatible with the geometry. \\
118 & Chemistry & \href{https://github.com/tblite/tblite}{\nolinkurl{tblite/tblite}} & PR \#326 / PR \#343 & GFN2-xTB multipole electrostatics, 3-D periodic Ewald decomposition, energy/force/stress consistency, unit cell translation and derivative cache & Different legitimate periodic unit cell representations of the same molecule should give consistent energies, forces, and stresses; multipolar Ewald's real/reciprocal paths and cached derivatives must be aligned. \\
119 & Chemistry & \href{https://github.com/MolSSI/QCEngine}{\nolinkurl{MolSSI/QCEngine}} & PR \#349 & Turbomole nprhessian/hessian selection, 3N\ensuremath{\times}3N Cartesian Hessian, projected/unprojected curvature, QCSchema result consistency & Turbomole Hessian analysis must recognize nprhessian/hessian records, infer the full 3N\ensuremath{\times}3N dimensions, and pass the unprojected Cartesian curvature to downstream reaction path/vibration analysis. \\
\end{longtable}
\endgroup

\end{document}